\documentclass[10pt,twocolumn,letterpaper]{article}

\usepackage{cvpr}
\usepackage{times}
\usepackage{epsfig}
\usepackage{graphicx}
\usepackage{epstopdf} 
\usepackage{amsmath}
\usepackage{amssymb}
\usepackage{algorithm}
\usepackage{algorithmicx}
\usepackage{algpseudocode}
\usepackage{cases}
\usepackage{subfigure}
\usepackage{overpic}
\usepackage{multirow}
\usepackage{color}
\usepackage{array}
\usepackage{tabularx}
\usepackage[pagebackref=true,breaklinks=true,letterpaper=true,colorlinks,bookmarks=false]{hyperref}

\cvprfinalcopy % *** Uncomment this line for the final submission

\def\cvprPaperID{477} % *** Enter the CVPR Paper ID here

\ifcvprfinal\fi
\begin{document}

%%%%%%%%% TITLE
\title{Re-ranking Person Re-identification with $k$-reciprocal Encoding}

\author{Zhun Zhong$^{\dag \ddag}$, Liang Zheng$^{\S}$, Donglin Cao$^{\dag \ddag}$, Shaozi Li$^{\dag \ddag}$\thanks{Corresponding author} \\
 \small{$^{\dag}$Cognitive Science Department, Xiamen University, China} \\
 \small{$^{\S}$University of Technology Sydney}\\ 
\small{$^{\ddag}$ Fujian Key Laboratory of Brain-inspired Computing 
Technique and Applications, Xiamen University} \\
{\tt\small \{zhunzhong007,liangzheng06\}@gmail.com \{another,szlig\}@xmu.edu.cn}
 }

\maketitle
\thispagestyle{empty}
%-------------------------------------------------------------------------
%%%%%%%%% ABSTRACT
\begin{abstract}
When considering person re-identification (re-ID) as a retrieval process, re-ranking is a critical step to improve its accuracy. Yet in the re-ID community, limited effort has been devoted to re-ranking, especially those fully automatic, unsupervised solutions. 
In this paper, we propose a $k$-reciprocal encoding method to re-rank the re-ID results. Our hypothesis is that if a gallery image is similar to the probe in the $k$-reciprocal nearest neighbors, it is more likely to be a true match.
Specifically, given an image, a $k$-reciprocal feature is calculated by encoding its $k$-reciprocal nearest neighbors into a single vector, which is used for re-ranking under the Jaccard distance.
The final distance is computed as the combination of the original distance and the Jaccard distance. Our re-ranking method does not require any human interaction or any labeled data, so it is applicable to large-scale datasets.
Experiments on the large-scale Market-1501, CUHK03, MARS, and PRW datasets confirm the effectiveness of our method\footnote{The source code is available upon request.}. 

\end{abstract}
%-------------------------------------------------------------------------
%%%%%%%%% BODY TEXT
\section{Introduction}
Person re-identification (re-ID) \cite{zheng2016personsurvery,bedagkar2014survey,liao2015lomo,martinel2016person, ma2015cross,ma2013domain} is a challenging task in computer vision. In general, re-ID can be regarded as a retrieval problem. Given a probe person, we want to search in the gallery for images containing the same person in a cross-camera mode. After an initial ranking list is obtained, a good practice consists of adding a re-ranking step, with the expectation that the relevant images will receive higher ranks.  In this paper, we thus focus on the re-ranking issue.

%----------figure k-nn-------------
\begin{figure}[!h]
\centering
\includegraphics[width=3.2in]{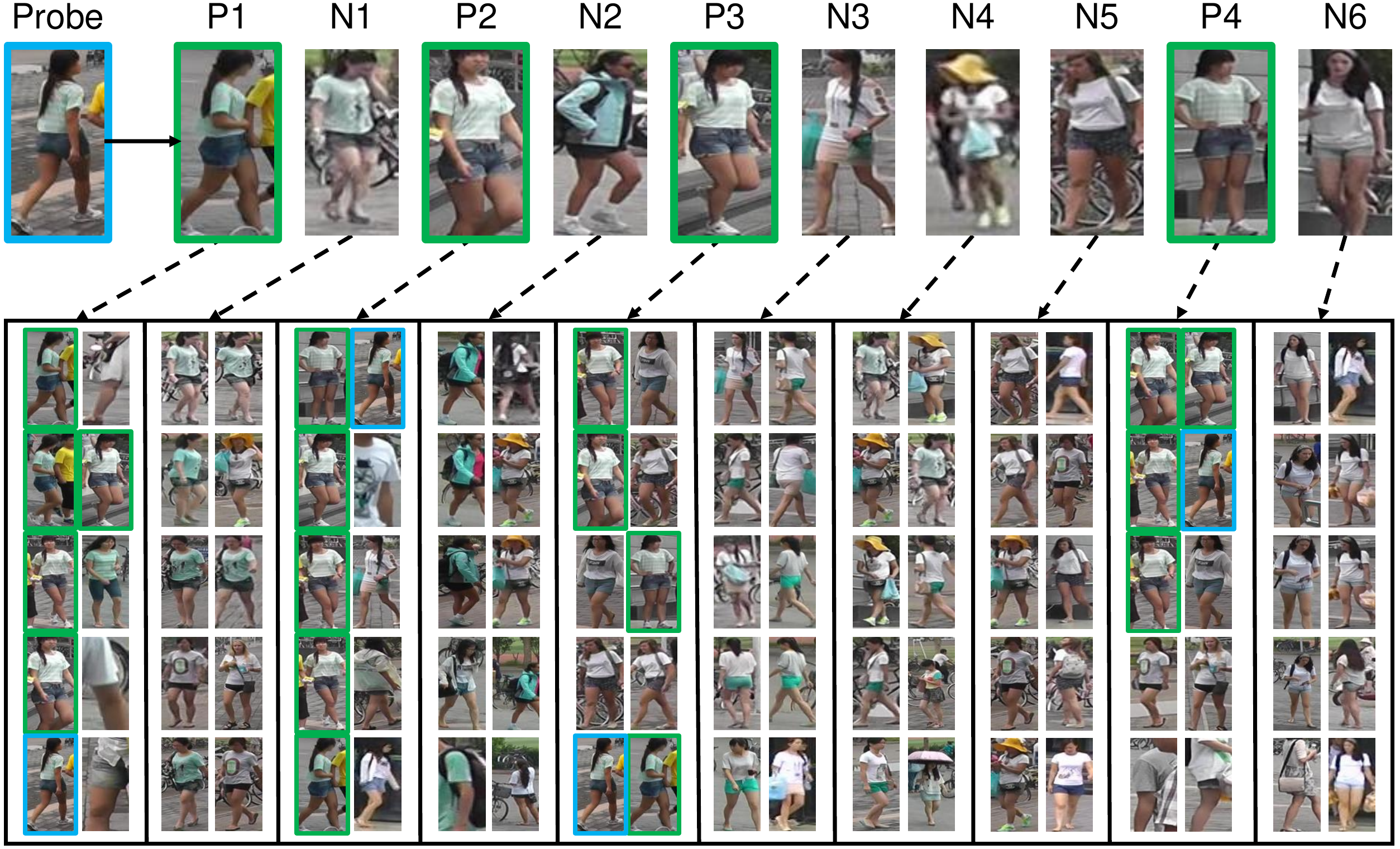}%
\caption{Illustration of the nearest neighborhoods of a person re-identification application. \textbf{Top:} The query and its 10-nearest neighbors, where P1-P4 are positives, N1-N6 are negatives. \textbf{Bottom:} Each two columns shows 10-nearest neighbors of the corresponding person. Blue and green box correspond to the probe and positives, respectively. We can observe that the probe person and positive persons are $10$-nearest neighbors reciprocally.
}
\label{fig:k-nn}
\end{figure}
%--------------------------------

Re-ranking has been mostly studied in generic instance retrieval \cite{chum2007total1, jegou2007contextual, qin2011hello,shen2012object-knn}. The main advantage of many re-ranking methods is that it can be implemented without requiring additional training samples, and that it can be applied to any initial ranking result. 

The effectiveness of re-ranking depends heavily on the quality of the initial ranking list. A number of previous works exploit the similarity relationships between top-ranked images (such as the $k$-nearest neighbors) in the initial ranking list \cite{chum2007total1,jegou2007contextual,qin2011hello,shen2012object-knn,ye2015ranking,yeperson-personre-2016}.
An underlying assumption is that if a returned image ranks within the $k$-nearest neighbors of the probe, it is likely to be a true match which can be used for the subsequent re-ranking. Nevertheless, situation may deviate from optimal cases: false matches may well be included in the $k$-nearest neighbors of the probe. 
For example, in Fig.~\ref{fig:k-nn}, P1, P2, P3 and P4 are four true matches to the probe, but all of them are not included in the top-4 ranks. We observe some false matches (N1-N6) receive high ranks. As a result, directly using the top-$k$ ranked images may introduce noise in the re-ranking systems and compromise the final result.

In literature, the $k$-reciprocal nearest neighbor \cite{jegou2007contextual, qin2011hello} is an effective solution to the above-mentioned problem, \ie, the pollution of false matches to the top-$k$ images. When two images are called $k$-reciprocal nearest neighbors, they are both ranked top-$k$ when the other image is taken as the probe. Therefore, the $k$-reciprocal nearest neighbor serves as a stricter rule whether two images are true matches or not. In Fig.~\ref{fig:k-nn}, we observe that the probe is a reciprocal neighbor to the true matched images, but not to the false matches. This observation identifies the true matches in the initial ranking list to improve the re-ranking results.

\begin{figure}[!t]
\centering
\includegraphics[width=1\linewidth]{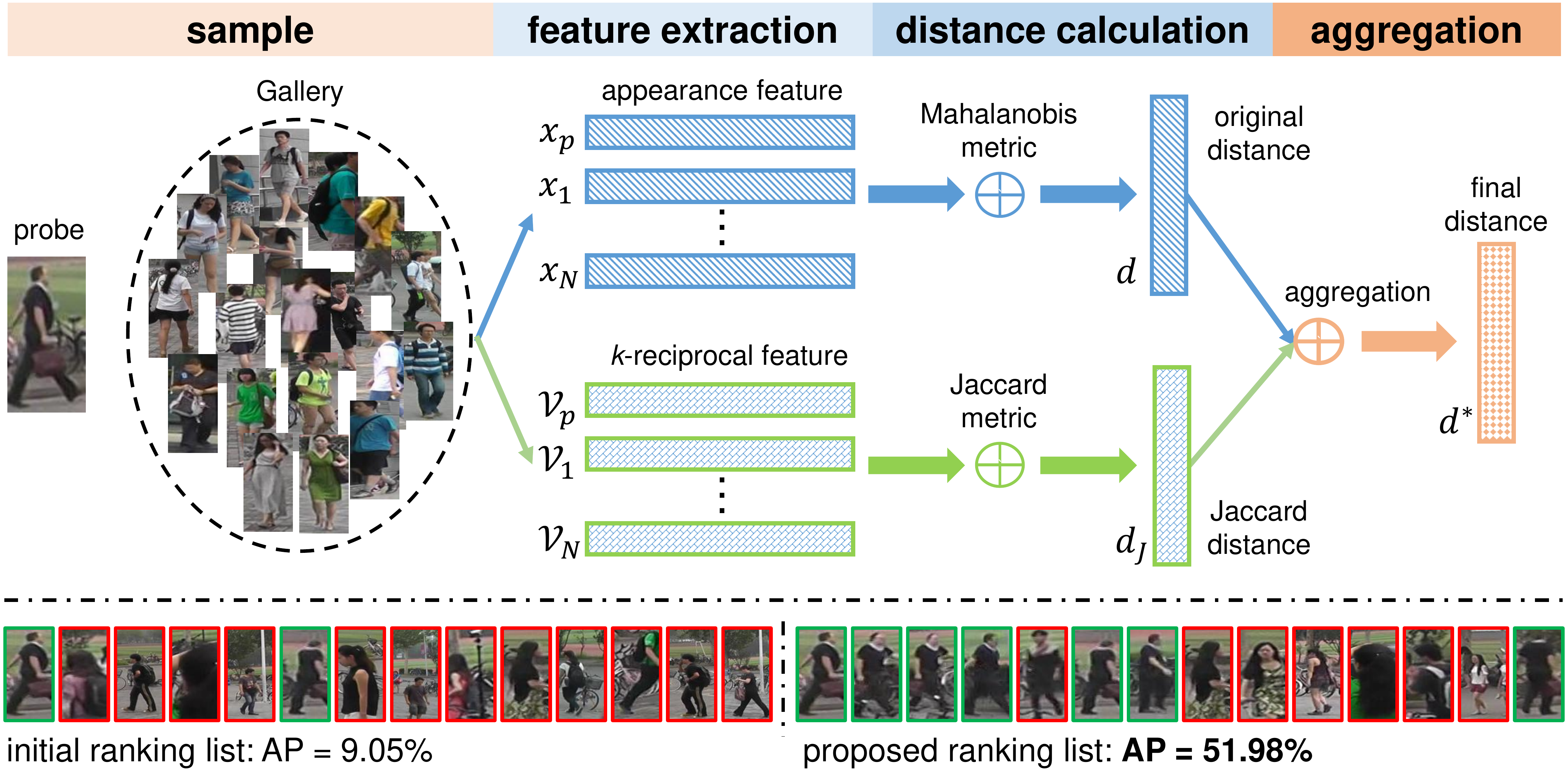}
\caption{Proposed re-ranking framework for person re-identification. Given a probe $p$ and a gallery, the appearance feature and $k$-reciprocal feature are extracted for each person. Then the original distance $d$ and Jaccard distance $d_J$ are calculated for each pair of the probe person and gallery person. The final distance $d^{\ast}$ is computed as the combination of $d$ and $d_J$, which is used to obtain the proposed ranking list.}
\label{fig:flow}
\end{figure}

Given the above considerations, this paper introduces a $k$-reciprocal encoding method for re-ID re-ranking. Our approach consists of three steps. First, we encode the weighted $k$-reciprocal neighbor set into a vector to form the $k$-reciprocal feature.  Then, the Jaccard distance between two images can be computed by their $k$-reciprocal features. Second, to obtain a more robust $k$-reciprocal feature, we develop a local query expansion approach to further improve the re-ID performance. Finally, the final distance is calculated as the weighted aggregation of the original distance and the Jaccard distance. It is subsequently used to acquire the re-ranking list. The framework of the proposed approach is illustrated in Fig.~\ref{fig:flow}.  To summarize, the contributions of this paper are:

\begin{itemize}
\item We propose a $k$-reciprocal feature by encoding the $k$-reciprocal feature into a singe vector. The re-ranking process can be easily performed by vector comparison.
\item Our approach does not require any human interaction or annotated data, and can be applied to any person re-ID ranking result in an automatic and unsupervised way.
\item The proposed method effectively improves the person re-ID performance on several datasets, including Market-1501, CUHK03, MARS, and PRW. In particular, we achieve the state-of-the-art accuracy on Market-1501 in both rank-1 and mAP.
\end{itemize}

%-------------------fig:flow-------------

\section{Related Work}
We refer the interested readers to  \cite{bedagkar2014survey, zheng2016personsurvery}  for a detailed review of person re-identification (re-ID). Here we focus on research that aims at re-ranking methods for object retrieval and particularly for re-ID.

\textbf{Re-ranking for object retrieval.} Re-ranking methods have been successfully studied to improve object retrieval accuracy \cite{zheng2016sift}. A number of  works utilize the $k$-nearest neighbors to explore similarity relationships to address the re-ranking problem.
Chum \emph{et  al.}  \cite{chum2007total1} propose the average query expansion (AQE) method, where a new query vector is obtained by averaging the vectors in the top-$k$ returned results, and is used to re-query the database.
To take advantage of the negative sample which is far away from the query image, Arandjelovi{\'c} and Zisserman \cite{arandjelovic2012three} develop the discriminative query expansion (DQE) to use a linear SVM to obtain a weight vector. The distance from the decision boundary is employed to revise the initial ranking list.
Shen \emph{et  al.} \cite{shen2012object-knn} make use of the \emph{k}-nearest neighbors of the initial ranking list as new queries to produce new ranking lists. The new score of each image is calculated depending on its positions in the produced ranking lists. 
More recently, sparse contextual activation (SCA) \cite{bai2016sparse} propose to encode the neighbor set into a vector, and to indicate samples similarity by generalized Jaccard distance.
To prevent the pollution of false matches to the top-$k$ images, the concept of $k$-reciprocal nearest neighbors is adopted in \cite{jegou2007contextual, qin2011hello}. 
In \cite{jegou2007contextual}, the contextual dissimilarity measure (CDM) is proposed to refine the similarity by iteratively regularizing the average distance of each point to its neighborhood. 
Qin \emph{et  al.} \cite{qin2011hello}  formally present the concept of $k$-reciprocal nearest neighbors. The k-reciprocal nearest neighbors are considered as highly relevant candidates, and used to construct closed set for re-ranking the rest of dataset.
Our work departs from both works in several aspects. We do not symmetrize nearest neighborhood relationship to refine the similarity as \cite{jegou2007contextual}, or directly consider the $k$-reciprocal nearest neighbors as top ranked samples like \cite{qin2011hello}. Instead we calculate a new distance between two images by comparing their $k$-reciprocal nearest neighbors.

\textbf{Re-ranking for re-ID.} Most existing person re-identification methods mainly focus on feature representation \cite{yang2014salient,gray2008viewpoint,liao2015lomo,zheng2015scalable,cnn-li2014deepreid} or metric learning \cite{liao2015lomo,kostinger2012KISSME, garcia2016modeling,martinel2015kernelized, zhang2016learningDNS}. Recently, several researchers \cite{garcia-niki-2017discriminant, nguyen2013re, ma2014query, zheng2015query,li2012common-person-couple,garcia2015personreranking,leng2015person-re-ranking, ye2015coupled-personre, yeperson-personre-2016} have paid attention to re-ranking based method in the re-ID community. 
Different from \cite{liu2013pop, wang2016human} and \cite{bai2017ssm}, which require human interaction or label supervision, we focus on an automatic and unsupervised solution.
Li \emph{et  al.} \cite{li2012common-person-couple} develop a re-ranking model by analyzing the relative information and direct information of near neighbors of each pair of images.
In \cite{garcia2015personreranking}, an unsupervised re-ranking model is learnt  by jointly considering the content and context information in the ranking list, which effectively remove ambiguous samples to improve the performance of re-ID.
Leng \emph{et  al.} \cite{leng2015person-re-ranking}  propose a bidirectional ranking method to revise the initial ranking list with the new similarity computed as the fusion of both content and contextual similarity.
Recently, the common nearest neighbors of different baseline methods are exploited to re-ranking task \cite{ye2015coupled-personre, yeperson-personre-2016}. 
Ye  \emph{et  al.} \cite{ye2015coupled-personre} combine the common nearest neighbors of global and local features as new queries, and 
 revise the initial ranking list by aggregating the new ranking lists of global and local features.
In \cite{yeperson-personre-2016}, the $k$-nearest neighbor set is utilized to calculate both similarity and dissimilarity from different baseline method, then the aggregation of similarity and dissimilarity is performed to optimize the initial ranking list.
Continues progress of these mentioned methods in re-ranking promises to make future contributions to discovering further information from $k$-nearest neighbors.
However, using the $k$-nearest neighbors to implement re-ranking directly may restrict the overall performance since false matches are often included.
To tackle this problem, in this paper, we investigate the importance of $k$-reciprocal neighbors in person re-ID and hence design a simple but effective re-ranking method.
%

%------------------------------------------------------------------------
\section{Proposed Approach}

\subsection{Problem Definition}

Given a probe person $p$ and the gallery set with $N$ images $\mathcal{G} = \{g_i\mid i = 1, 2, ... N\}$, the original distance between two persons $p$ and $g_i$ can be measured by Mahalanobis distance,
    \begin{eqnarray}
          d(p,g_i)=(x_p- x_{g_i})^\top  \mathbf{M}( x_p - x_{g_i})
    \end{eqnarray}
where $x_p$ and $x_{g_i}$ represents the appearance feature of probe $p$ and gallery $g_i$, respectively, and $\mathbf{M}$ is a positive semidefinite matrix.

The initial ranking list $\mathcal{L}(p, \mathcal{G}) = \{g_1^0, g_2^0, ... g_N^0\}$ can be obtained according to the pairwise original distance between probe $p$ and gallery $g_i$, where  $d(p,g_i^0) <  d(p,g_{i+1}^0)$.
Our goal is to re-rank $\mathcal{L}(p, \mathcal{G})$, so that more positive samples rank top in the list, and thus to improve the performance of person re-identification (re-ID).

\subsection{$K$-reciprocal Nearest Neighbors}
Following \cite{qin2011hello}, we define $N(p,k)$ as the $k$-nearest neighbors (\emph{i.e.} the top-$k$  samples of the ranking list) of a probe $p$:
 % \begin{center}
    \begin{eqnarray}
   N(p,k) = \{g_1^0, g_2^0, ...,  g_k^0\}, \left | N(p,k) \right |  = k 
\end{eqnarray}
  %\end{center}
 %
 where $\left|\cdot \right|$ denotes the number of candidates in the set. The $k$-reciprocal nearest neighbors $\mathcal{R}(p,k)$ can be defined as,
  %  \begin{center}
    \begin{eqnarray}
   \mathcal{R}(p,k) = \{g_i\mid (g_i \in N(p,k)) \wedge (p \in N(g_i,k))\} 
   \label{equ:k-re}
\end{eqnarray}
 % \end{center}
%
According to the previous description,  the $k$-reciprocal nearest neighbors are more related to probe $p$ than $k$-nearest neighbors.
However, due to variations in illuminations, poses, views and occlusions, the positive images may be excluded from the $k$-nearest neighbors, and subsequently not be included in the $k$-reciprocal nearest neighbors.
To address this problem, we incrementally add the $\frac{1}{2}k$-reciprocal nearest neighbors of each candidate in $\mathcal{R}(p,k)$ into a more robust set $\mathcal{R}^{\ast}(p,k)$ according to the following condition
    \begin{eqnarray}
    \begin{array}{l}
\mathcal{R}^{\ast}(p,k) \gets \mathcal{R}(p,k) \cup \mathcal{R}(q,\frac{1}{2}k)  \\ \\
\emph{s.t.} \left | \mathcal{R}(p,k) \cap \mathcal{R}(q,\frac{1}{2}k) \right |  \geqslant \frac{2}{3} \left |  \mathcal{R}(q,\frac{1}{2}k) \right |, \\ \\
\forall  q \in \mathcal{R}(p,k)
   \label{equ:k-re-expan}
   \end{array}
\end{eqnarray}
By this operation, we can add into $\mathcal{R}^{\ast}(p,k)$ more positive samples which are more similar to the candidates in $\mathcal{R}(p,k)$ than to the probe $p$. This is stricter against including too many negative samples compared to \cite{qin2011hello}.
%---------------------------------------------------------------
\begin{figure}[!t]
\centering
\includegraphics[width=0.7\linewidth]{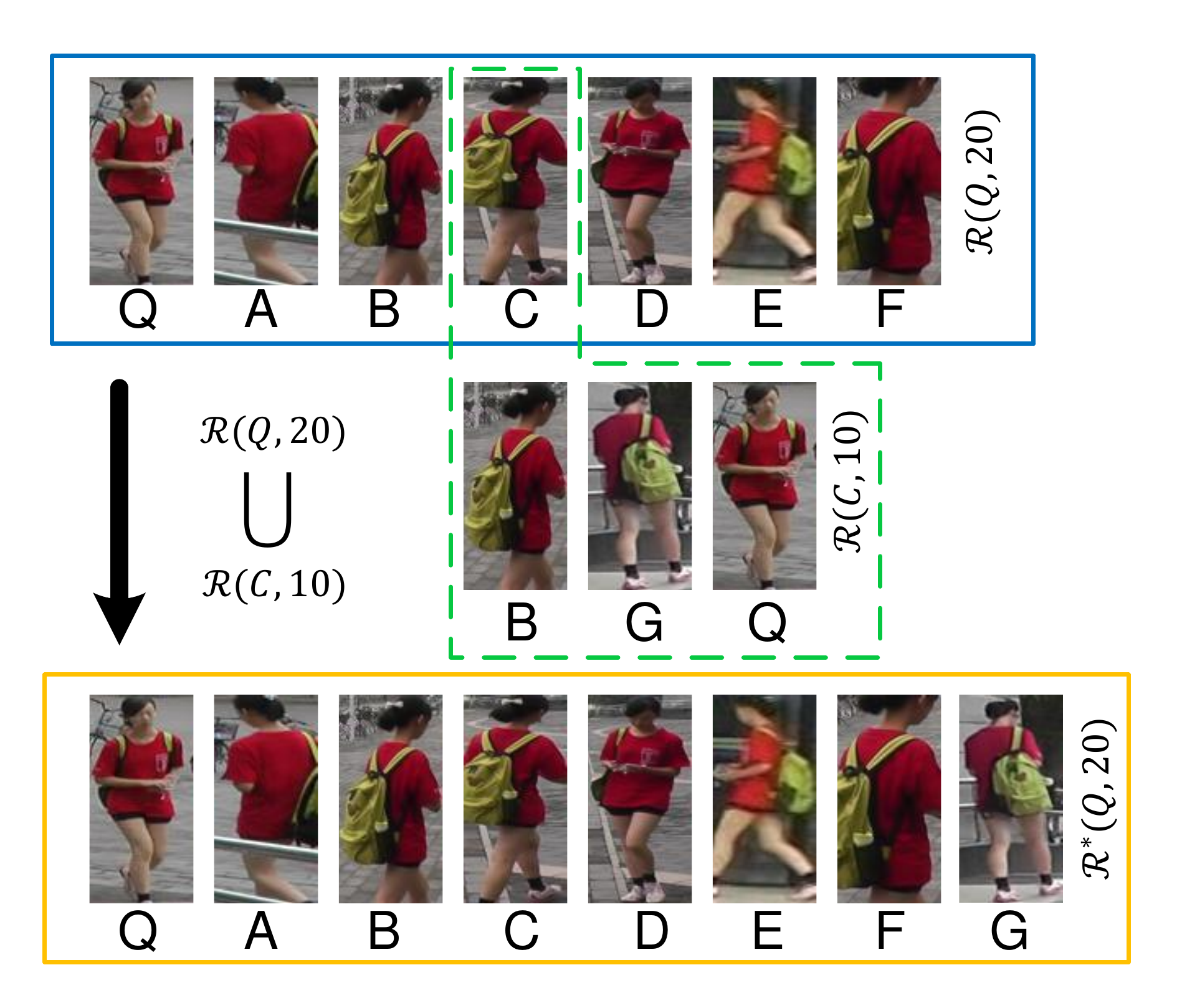}%
\caption{Example of the k-reciprocal neighbors expansion process. The positive person G which is similar to C is added into  $\mathcal{R}^{\ast}(Q,20)$.}
\label{fig:k-reciprocal-step}
\end{figure}
%--------------------------------
%-------------------------------------------------------------------------
%---------------------------------------------------------------
% \floatname{algorithm}{Algorithm}
% \renewcommand{\algorithmicrequire}{\textbf{Input:}}
% \renewcommand{\algorithmicensure}{\textbf{Output:}}
%     \begin{algorithm}
%        \caption{:Expansion process}
%         \begin{algorithmic}[1] 
%             \Require Initial $k$-reciprocal nearest neighbor set $\mathcal{R}(p,k)$
%             \Ensure Expanded $\mathcal{R}^{\ast}(p,k)$
%                 \State $\mathcal{R}^{\ast}(p,k) \gets \mathcal{R}(p,k)$
%                 \For{$i \in \mathcal{R}(p,k)$}
%                 \If {$\left | \mathcal{R}(p,k) \cap \mathcal{R}(i,\frac{1}{2}k) \right |  \geqslant \frac{2}{3} \left |  \mathcal{R}(i,\frac{1}{2}k) \right | $}
%                     \State $\mathcal{R}^{\ast}(p,k) \gets \mathcal{R}(p,k) \cup \mathcal{R}(i,\frac{1}{2}k)$
%                 \EndIf 
%                 \EndFor 
%                 \State \Return{$\mathcal{R}^{\ast}(p,k)$}
%         \end{algorithmic}
%          \label{algorithm 1}
%     \end{algorithm}
%---------------------------------------------------------------
%
In Fig.~\ref{fig:k-reciprocal-step}, we show an example of the expansion process. 
Initially, the hard positive G is missed out in $\mathcal{R}(Q,20)$.
Interestingly, G is included in $\mathcal{R}(C,10)$, which is beneficial information for bringing positive G back.
Then, we can apply Eq. \ref{equ:k-re-expan} to add G into $\mathcal{R}^{\ast}(Q,20)$.
Therefore, after expansion process, more positive samples could be  added into $\mathcal{R}^{\ast}(p,k)$.
Different from \cite{qin2011hello}, we do not directly take the candidates in $\mathcal{R}^{\ast}(p,k)$ as top ranked images. 
Instead, we consider $\mathcal{R}^{\ast}(p,k)$ as contextual knowledge to re-calculate the distance between the probe and gallery. 

\subsection{Jaccard Distance}

In this subsection, we re-calculate the pairwise distance between the probe $p$ and the gallery $g_i$ by comparing their $k$-reciprocal nearest neighbor set.
As described earlier  \cite{bai2016sparse} \cite{yeperson-personre-2016}, we believe that if two images are similar, their k-reciprocal nearest neighbor sets overlap, \emph{i.e.}, there are some duplicate samples in the sets.
And the more duplicate samples, the more similar the two images are.
The new distance between  $p$  and $g_i$ can be calculated by  the Jaccard metric of their $k$-reciprocal sets as:
    \begin{eqnarray}
   d_J(p,{g_i})= 1 - \frac{\left | \mathcal{R}^{\ast}(p,k) \cap \mathcal{R}^{\ast}(g_i,k) \right |}{\left | \mathcal{R}^{\ast}(p,k) \cup \mathcal{R}^{\ast}(g_i,k) \right |}
   \label{jaccard}
\end{eqnarray}
  where $\left|\cdot \right|$ denotes the number of candidates in the set.  We adopt Jaccard distance to name this new distance. 
%  
%After contextual distance  calculated, we can get the re-ranked list by descending sort.
Although the above method could capture the similarity relationships between two images, there still remains three obvious shortcomings:

\begin{itemize}
\item It is very time-consuming to get the intersection and union of two neighbor sets  $\mathcal{R}^{\ast}(p,k)$ and  $\mathcal{R}^{\ast}(g_i,k)$ in many cases, and it becomes more challenging while the Jaccard distance is needed to be calculated for all image pairs. An alternative way is to encode the neighbor set into an easier but equivalent vector, reducing the computational complexity greatly, while maintaining original structure in neighbor set.
\item The distance calculation method weighs all neighbors equally, leading to a simple but not discriminative neighbor set. In fact, neighbors that are closer to probe $p$ are more likely to be true positives. Therefore, it is convincing and reasonable to re-calculate weights based on the original distance, and assign large weights to ’nearer’ samples.
\item Simply taking the contextual information into account will pose considerable barriers when attempting to measure similarity between two persons, since unavoidable variation makes it difficult to discriminate sufficient contextual information. Hence, incorporating original distance and Jaccard distance becomes important for a robust distance.
\end{itemize}

%
%-------------------------------------------------------------------------

Inspired by \cite{bai2016sparse}, the $k$-reciprocal feature is proposed to address the first two shortcomings, by encoding the $k$-reciprocal nearest neighbor set into a vector $\mathcal{V}_p = \left[\mathcal{V}_{p, g_1}, \mathcal{V}_{p, g_2}, ..., \mathcal{V}_{p, g_N} \right]$, where $\mathcal{V}_{p, g_i}$ is initially defined by a binary indicator function  as
\begin{eqnarray}
\mathcal{V}_{p, g_i} =
\begin{cases}
1& \text{ if }  g_i \in \mathcal{R}^{\ast}(p,k)   \\ 
0& \text{ otherwise. } 
\end{cases} 
\label{value}
\end{eqnarray}
In this way, the $k$-reciprocal neighbor set can be represented as an $N$-dimensional vector, with each item of the vector indicating whether the corresponding image is included in $\mathcal{R}^{\ast}(p,k)$. 
However, this function still consider each neighbor as equal. Intuitively, the neighbor who is closer to the probe $p$ should be more similar with the probe $p$. 
Thus, we reassign weights according to the original distance between the probe and its neighbor, we redefine Eq.~\ref{value} by the Gaussian kernel of the pairwise distance as
\begin{eqnarray}
\mathcal{V}_{p, g_i} =
\begin{cases}
\mathrm{e}^{-d(p, g_i)} & \text{if } g_i \in \mathcal{R}^{\ast}(p,k)   \\ 
0 & \text{otherwise}.
\end{cases} 
\label{re-value}
\end{eqnarray}
In this way, the hard  weighting (0 or 1) is converted into soft weighting, with closer neighbors assigned larger weights while farther neighbors smaller weights.
Based on the above definition, the number of candidates in the intersection and union set can be calculated as
 %   \begin{center}
    \begin{eqnarray}
 \left| \mathcal{R}^{\ast}(p,k) \cap \mathcal{R}^{\ast}(g_i,k) \right | = {\left \|\min(\mathcal{V}_{p}, \mathcal{V}_{g_i})  \right \|}_1 \\
 \left | \mathcal{R}^{\ast}(p,k) \cup \mathcal{R}^{\ast}(g_i,k) \right | = {\left \|\max (\mathcal{V}_{p}, \mathcal{V}_{g_i}) \right \|}_1
\end{eqnarray}
%  \end{center}
where $\min$ and $\max$ operate the element-based minimization and maximization for two input vectors.
${\left \| \cdot \right \|}_1$ is $L_1$ norm. Thus we can rewrite the Jaccard distance in Eq.~\ref{jaccard} as
 %  \begin{center}
    \begin{eqnarray}
   d_J(p,{g_i})= 1 - \frac{\sum_{j=1}^{N}\min (\mathcal{V}_{p,g_j}, \mathcal{V}_{g_i,g_j})}{\sum_{j=1}^{N}\max (\mathcal{V}_{p,g_j}, \mathcal{V}_{g_i,g_j})}
   \label{re-jaccard}
\end{eqnarray}
 % \end{center}
By formula transformation from Eq.~\ref{jaccard} to Eq.~\ref{re-jaccard}, we have succeed in converting the set comparison problem into pure vector calculation, which is much easier practically.

%-------------------------------------------------------------------------
\subsection{Local Query Expansion}
Emulating the idea that the images from the same class may share similar features, we use the $k$-nearest neighbors of the probe $p$ to implement the local query expansion. 
 The local query expansion is defined as
% \begin{center}
\begin{eqnarray}
\mathcal{V}_{p} = {\frac{1}{\left| N(p,k) \right|}}   \underset{g_i \in N(p,k)}{\sum} \mathcal{V}_{g_i}
\label{local-query}
\end{eqnarray}
%\end{center}
As a result, the $k$-reciprocal feature $\mathcal{V}_{p}$ is expanded by the $k$-nearest neighbors of probe $p$. Note that, we implement this query expansion both on the probe $p$ and galleries $g_i$. 
Since there will be noise in the $k$-nearest neighbors, we limit the size of $N(p,k)$ used in the local query expansion to a smaller value.
%than that used in the weighting contextual similarity. 
%
In order to distinguish between the size of $\mathcal{R}^{\ast}(g_i,k)$ and $N(p,k)$ used in Eq.~\ref{re-value} and Eq.~\ref{local-query}, we denote the former as $k_1$ and the latter as $k_2$, respectively, where $k_1>k_2$ .
%-------------------------------------------------------------------------
\subsection{Final Distance}
In this subsection, we focus on the third shortcoming of Eq. \ref{jaccard}.
While most existing  re-ranking methods ignore the importance of original distance in re-ranking, we jointly aggregate the original distance and Jaccard distance to revise the initial ranking list, the final distance $d^{\ast}$ is defined as
 % \begin{center}
\begin{eqnarray}
d^{\ast}(p, g_i) = (1 - \lambda) d_J(p,{g_i}) +  \lambda{d(p, g_i)}
\label{final-d}
\end{eqnarray}
%\end{center}
where $\lambda \in \left[0, 1\right]$ denotes the penalty factor, it penalizes galleries far away from the probe $p$. When $\lambda = 0$, only the k-reciprocal distance is considered. On the contrary, when $\lambda$ = 1, only the original distance is considered. The effect of $\lambda$ is discussed in section \ref{section-ex}. Finally, the revised ranking list $\mathcal{L}^{\ast}(p, \mathcal{G})$ can be obtained by ascending sort of the final distance.

\subsection{Complexity Analysis}
In the proposed method, most of the computation costs focus on pairwise distance computing for all gallery pairs. Suppose the size of the gallery set is $N$, the computation complexity required for the distance measure and the ranking process is $O(N^2)$ and $O(N^2logN)$, respectively. However, in practical applications, we can calculate the pairwise distance and obtain the ranking lists for the gallery in advance offline. As a result, given a new probe $p$, we only need to compute the pairwise distance between $p$ and gallery with computation complexity $O(N)$ and to rank all final distance with computation complexity $O(NlogN)$.

%-------------------------------------------------------------------------
\section{Experiments}
\label{section-ex}

\subsection{Datasets and Settings}
 \textbf{Datasets} Because our re-ranking approach is based on the comparison of similar neighbors between two persons, we conducted experiments on four large-scale person re-identification (re-ID) benchmark datasets that contain multiple positive samples for each probe in the gallery : including two image-based datasets, Market-1501 \cite{zheng2015scalable}, CUHK03 \cite{CUHK03} , a video-based dataset MARS \cite{zheng2016mars}, and an end-to-end dataset PRW \cite{prw} (see Table~\ref{tabel:dataset} for an overview).

 \textbf{Market-1501} \cite{zheng2015scalable} is currently the largest image-based re-ID benchmark dataset. It contains 32,668 labeled bounding boxes of 1,501 identities captured from 6 different view points. The bounding boxes are detected using Deformable Part Model (DPM) \cite{dpm}. The dataset is split into two parts: 12,936 images with 751 identities for training and 19,732 images with 750 identities for testing. In testing, 3,368 hand-drawn images with 750 identities are used as probe set to identify the correct identities on the testing set. We report the single-query evaluation results for this dataset.

 \textbf{CUHK03} \cite{CUHK03} contains 14,096 images of 1,467 identities. Each identity is captured from two cameras in the CUHK campus, and has an average of 4.8 images in each camera. The dataset provides both manually labeled bounding boxes and DPM-detected bounding boxes. In this paper, both experimental results on `labeled' and `detected' data are presented.

 \textbf{MARS} \cite{zheng2016mars} is the largest video-based re-ID benchmark dataset to date, containing 1,261 identities and around 20,000 video sequences. These sequences are collected from 6 different cameras and  each identity has 13.2 sequences on average. Each sequence is automatically obtained  by the DPM as pedestrian detector and the GMMCP \cite{dehghan2015gmmcp} as tracker. In addition, the dataset also contains 3,248 distractor sequences. The dataset is fixedly split into training and test sets, with 631 and 630 identities, respectively. In testing, 2,009 probes are selected for query.

 \textbf{PRW} \cite{prw} is an end-to-end large-scale dataset. It is composed of 11,816 frames of 932 identities captured from six different cameras. A total of 43,110 annotated person bounding boxes are  generated from these frames. Given a query bounding box, the dataset aims to first perform pedestrian detection on the raw frames to generate the gallery, and identify the correct bounding boxes from the gallery. The dataset is divided into a training set with 5,704 frames of 482 identities
and a test set with 6,112 frames of 450 identities. In testing, 2,057 query images for 450 identities are selected for evaluation. A detected bounding box is considered correct if its IoU value with the ground truth is above 0.5.

%-----------------------------------------------------------------
\begin{table}
\footnotesize
\begin{center}
\caption{\label{tabel:dataset} The details of datasets used in our experiments.}\vspace{.03in}
\newcolumntype{C}{>{\centering\arraybackslash}X}%
\newcolumntype{R}{>{\raggedleft\arraybackslash}X}%
\begin{tabularx}{\linewidth}{ l|C|C|C|C }
\hline
Datasets & \# ID & \# box &  \# box/ID  & \# cam\\
%\hline
\hline
Market-1501 \cite{zheng2015scalable} &1,501 & 32,643 & 19.9 & 6 \\
%\hline
CUHK03 \cite{CUHK03} & 1,467 & 14,096 & 9.7 & 2 \\
%\hline
MARS \cite{zheng2016mars} & 1,261 & 1,067,516 & 13.2 & 6 \\
%\hline
PRW \cite{prw} & 932 & 34,304 & 36.8 & 6 \\
\hline
\end{tabularx}
\end{center}
\vspace{-.2in}
\end{table}
%----------------------------------------------------------------------------

 \textbf{Evaluation metrics} We use two evaluation metrics to evaluate the performance of re-ID methods on all datasets. The first one is the Cumulated Matching Characteristics (CMC).  Considering re-ID as a ranking problem, we report the cumulated matching accuracy at rank-1. The other one is the mean average precision (mAP) considering re-ID as an object retrieval problem,  as described in \cite{zheng2015scalable}. 

 \textbf{Feature representations} The Local Maximal Occurrence (LOMO) features are used to represent the person appearance \cite{liao2015lomo}. It is robust to view changes and illumination variations. In addition, the ID-discriminative Embedding (IDE) feature proposed in \cite{prw} is used. The IDE extractor is effectively  trained on classification model including CaffeNet \cite{krizhevsky2012imagenet} and ResNet-50 \cite{He_2016_CVPR_resnet}. It generates a 1,024-dim  (or 2,048-dim) vector for each image, which is effective in large-scale re-ID datasets. For the convenience of description, we abbreviate the IDE trained on CaffeNet and ResNet-50 to IDE (C) and IDE (R) respectively.  We use these two methods as the baseline of our re-ID framework.

%-----------------------------------------------------------------
\begin{table}
\footnotesize
\begin{center}
\caption{\label{tabel:Market-compare} Comparison among various methods with our re-ranking approach on the  Market-1501 dataset.}\vspace{.03in}
\newcolumntype{C}{>{\centering\arraybackslash}X}%
\newcolumntype{R}{>{\raggedleft\arraybackslash}X}%
\begin{tabularx}{\linewidth}{ l|C|C }
%\begin{tabular}{|l|c|c|}
\hline
Method & Rank 1 & mAP \\
\hline
\hline
BOW \cite{zheng2015scalable}  & 35.84 & 14.75 \\ 
BOW + Ours  & 39.85	 & 19.90  \\
BOW + KISSME & 42.90	&19.41 \\
BOW  + KISSME + Ours & 44.77 &25.64 \\
BOW + XQDA & 41.39 & 19.72 \\
BOW   + XQDA + Ours & 42.61	 & 24.98  \\
\hline
LOMO + KISSME & 41.12	 & 19.02 \\
LOMO  + KISSME + Ours & 45.22  & 28.44  \\
LOMO + XQDA \cite{liao2015lomo} & 43.56	 & 22.44  \\
LOMO   + XQDA + Ours & 48.34	& 32.21  \\
\hline
IDE (C) \cite{prw}  & 55.87	 & 31.34  \\
IDE (C) + AQE \cite{chum2007total1} & 57.69 & 35.25  \\
IDE (C) + CDM \cite{jegou2007contextual} & 58.02 & 34.54 \\
IDE (C)  + Ours&  58.79  & 42.06   \\
IDE (C) + XQDA  & 57.72	 & 35.95  \\
IDE (C) + XQDA + Ours&  61.25  & 46.79  \\
IDE (C) + KISSME  & 58.61	 & 35.40  \\
IDE (C) + KISSME + Ours&  61.82  & 46.81 \\
\hline
IDE (R) \cite{prw} & 72.54	 & 46.00  \\
IDE (R) + AQE \cite{chum2007total1} & 73.20 & 50.14  \\
IDE (R) + CDM \cite{jegou2007contextual} & 73.66 & 49.53  \\
IDE (R) + Ours &  74.85 & 59.87  \\
IDE (R) + XQDA  & 71.41	 & 48.89  \\
IDE (R) + XQDA + Ours& 75.14  & 61.87 \\
IDE (R) + KISSME  & 73.60	 & 49.05  \\
\textbf{IDE (R) + KISSME + Ours}&  \textbf{77.11} & \textbf{63.63} \\
\hline
\end{tabularx}
\end{center}
\vspace{-.2in}
\end{table}
%----------------------------------------------------------------------------

%------------------------------------------------------------------------
\subsection{Experiments on Market-1501}
We first evaluate our method on the largest image-based re-ID dataset. In this dataset, in addition to using LOMO and IDE features, we also use  the BOW \cite{zheng2015scalable} feature. We trained the IDE feature on CaffeNet \cite{krizhevsky2012imagenet} and ResNet-50 \cite{He_2016_CVPR_resnet}.  We set $k_1$ to 20, $k_2$ to 6, and $\lambda$ to 0.3. Results among various methods with our method are shown in Table~\ref{tabel:Market-compare}. Our method consistently improves the rank-1 accuracy and mAP with all features, even with the IDE (R) which is trained on the powerful ResNet-50 model. 
Our method gains $3.06\%$ improvement in rank-1 accuracy and  significant $13.99\%$ improvement in mAP for IDE (R). Moreover, experiments conducted with two metrics, KISSME \cite{kostinger2012KISSME} and XQDA \cite{liao2015lomo} verify the effectiveness of our method on different distance metrics.
Comparing with two popular re-ranking methods, average query expansion (AQE) \cite{chum2007total1} and contextual dissimilarity measure (CDM) \cite{jegou2007contextual}, our method outperforms them both in rank-1 accuracy and mAP.  Many existing re-ranking methods of person re-id are for single-shot setting or require human interaction \cite{liu2013pop,wang2016human}. Therefore, these methods are not directly comparable to our method. 

Table~\ref{tabel:Market-state} compares the performance of our best approach, IDE (R) + KISSME + ours, with other state-of-the-art methods. Our best method impressively outperforms the previous work and achieves large margin advances compared with the state-of-the-art results in rank-1 accuracy, particularly in mAP.

%-----------------------------------------------------------------
\begin{table}
\footnotesize
\begin{center}
\caption{\label{tabel:Market-state} Comparison of our method with state-of-the-art on the Market-1501 dataset. %SQ denotes the single query, and MQ denotes the multi-query. 
%Our method outperforms the previous work and advances the state-of-the-art results using single query.
}\vspace{.03in}
\newcolumntype{C}{>{\centering\arraybackslash}X}%
\newcolumntype{R}{>{\raggedleft\arraybackslash}X}%
\begin{tabularx}{\linewidth}{ l|C|C }
%\begin{tabular}{|l|c|c|}
\hline
Method & Rank 1 & mAP \\
\hline
\hline
SDALF \cite{2010sdalf} & 20.53 & 8.20 \\
eSDC \cite{zhao2013ESDC}  & 33.54 & 13.54\\
BOW  \cite{zheng2015scalable}  & 34.40 & 14.09 \\
PersonNet \cite{wu2016personnet} & 37.21 & 18.57\\
dCNN \cite{su2016deepcann} & 39.40 & 19.60\\
LOMO + XQDA \cite{liao2015lomo} & 43.79 & 22.22 \\
MSTCNN \cite{liu2016multi} & 45.10 & -\\
WARCA \cite{jose2016WARCA}& 45.16 & - \\
MBCNN \cite{ustinova2015MBCNN}   & 45.58 & 26.11 \\
HistLBP+kLFDA \cite{karanam2016comprehensive} & 46.50 & - \\
TMA \cite{martinel2016temporal}  & 47.92 & 22.31 \\
DLDA \cite{wu2016deep} & 48.15 & 29.94 \\
CAN \cite{liu2016end}  & 48.24 & 24.43 \\
SCSP \cite{chen2016similarity} & 51.90 & 26.35 \\
DNS \cite{zhang2016learningDNS}  & 61.02 & 35.68 \\
Gated \cite{varior2016gated}  & 65.88 & 39.55 \\
\hline
% BOW \cite{zheng2015scalable} - (MQ)  & 42.14 & 19.20 \\
% BOW + HS \cite{zheng2015scalable} - (MQ)  & 47.25 & 21.88 \\
% LOMO + XQDA \cite{liao2015lomo} - (MQ) & 54.13 & 28.41 \\
% S-LSTM  \cite{varior2016siameseLSTM} -(MQ) & 61.60 & 35.31 \\
% DNS \cite{zhang2016learningDNS} - (MQ) & 71.56 & 46.03 \\
% Gated \cite{varior2016gated} - (MQ) & 76.04 & 48.45 \\
% \hline
\textbf{IDE (R) + KISSME + Ours}&  \textbf{77.11} & \textbf{63.63} \\
\hline
\end{tabularx}
\end{center}
\vspace{-.1in}
\end{table}
%----------------------------------------------------------------------------

%------------------------------------------------------------------------
\subsection{Experiments on CUHK03}

Following the single-shot setting protocol in \cite{CUHK03}, we split the dataset into a training set containing 1,160 identities and a testing set containing 100 identities. The test process is repeated with 20 random splits. We set $k_1$ to 7, $k_2$ to 3 , and $\lambda$ to 0.85. Results for single-shot setting are shown in Table~\ref{tabel:CUHK03-compare-single}. 
As we can see that, when using IDE feature, our re-ranking results are almost equivalent to raw results.
It is reasonable that our approach does not work. Since there is only one positive for each identity in the gallery, our approach could not obtain sufficient contextual information.
Even so, our approach gains nearly 1\% improvement for rank-1 accuracy and mAP while applying LOMO feature on both `labeled' and `detected' setting, except LOMO + XQDA in `labeled' setting. Experiments show that, in the case of single-shot setting, our method does no harm to results, and has the chance to improve the performance.
%

%-----------------------------------------------------------------

\begin{table}
\footnotesize
\begin{center}
\caption{\label{tabel:CUHK03-compare-single} Comparison among various methods with our re-ranking approach on the  CUHK03 dataset under the single-shot setting.}\vspace{.03in}
%\begin{tabular}{|l|c|c|c|c|}
\newcolumntype{C}{>{\centering\arraybackslash}X}%
\newcolumntype{R}{>{\raggedleft\arraybackslash}X}%
\begin{tabularx}{\linewidth}{ l|C|C|C|C }
\hline
\multirow{2}{*}{Method} & \multicolumn{2}{c|}{Labeled} & \multicolumn{2}{c}{Detected} \\
\cline{2-5}
& Rank 1 &  mAP  & Rank 1 & mAP \\
\hline
\hline
LOMO + XQDA \cite{liao2015lomo} & 49.7  & 56.4 & 44.6 & 51.5 \\ 
LOMO + XQDA + Ours  & 50.0  & 56.8 & 45.9  & 52.6  \\
\hline
IDE (C) \cite{prw} & 57.0 & 63.1 & 54.1	 & 60.4 \\
IDE (C)  + Ours&  57.2  & 63.2 &  54.2  & 60.5 \\
IDE (C) + XQDA  &  61.7 & 67.6& 58.9 & 64.9\\
IDE (C)  + XQDA + Ours&  61.6 & 67.6 &  58.5 & 64.7 \\
\hline
\end{tabularx}
\end{center}
\vspace{-.1in}
\end{table}
%----------------------------------------------------------------------------

Apart from the previous evaluation method, we also report results using a new training/testing protocol similar to that of Market-1501.
The new protocol splits the dataset into training set and testing set, which consist of 767 identities and 700 identities respectively. In testing, we randomly select one image from each camera as the query for each identity and use the rest of images to construct the gallery set. The new protocol has two advantages:1) For each identity, there are multiple ground truths in the gallery. This is more consistent with practical application scenario. 2) Evenly dividing the dataset into training set and testing set at once helps avoid repeating training and testing multiple times. 
The divided training/testing sets and the evaluation code are available in our source code.
We set $k_1$ to 20, $k_2$ to 6, and $\lambda$ to 0.3. Results in Table~\ref{tabel:CUHK03-compare-multi} show that, in all cases, our method significantly improves rank-1 accuracy and mAP. Especially for IDE(R) + XQDA, our method gains an increase of 6.1\% in rank-1 accuracy and 10.7\% in mAP on `labeled' setting.

%-----------------------------------------------------------------

\begin{table}
\footnotesize
\begin{center}
\caption{\label{tabel:CUHK03-compare-multi} Comparison among various methods with our re-ranking approach on the CUHK03 dataset under the new training/testing protocol.}\vspace{.03in}
\newcolumntype{C}{>{\centering\arraybackslash}X}%
\newcolumntype{R}{>{\raggedleft\arraybackslash}X}%
\begin{tabularx}{\linewidth}{ l|C|C|C|C }
%\begin{tabular}{|l|c|c|c|c|}
\hline
\multirow{2}{*}{Method} & \multicolumn{2}{c|}{Labeled} & \multicolumn{2}{c}{Detected} \\
\cline{2-5}
  & Rank 1 &  mAP  & Rank 1 & mAP \\
\hline
\hline
LOMO + XQDA \cite{liao2015lomo} & 14.8 & 13.6 & 12.8 & 11.5  \\ 
LOMO + XQDA + Ours  & 19.1 & 20.8 & 16.6 & 17.8  \\
\hline
IDE (C) \cite{prw} & 15.6	& 14.9 & 15.1 & 14.2\\
IDE (C)   + Ours& 19.1	& 21.3 & 19.3 & 20.6 \\
IDE (C)  + XQDA  & 21.9	& 20.0 & 21.1 & 19.0 \\
IDE (C)   + XQDA + Ours& 25.9 & 27.8 & 26.4 & 26.9\\
\hline
IDE (R)  \cite{prw}  & 22.2	& 21.0 & 21.3 & 19.7 \\
IDE (R) + Ours & 26.6 & 28.9 & 24.9 & 27.3\\
IDE (R)  + XQDA  & 32.0	& 29.6 & 31.1 & 28.2 \\
\textbf{IDE (R)   + XQDA + Ours}& \textbf{38.1} & \textbf{40.3} & \textbf{34.7} & \textbf{37.4}\\
\hline
\end{tabularx}
\end{center}
\vspace{-.1in}
\end{table}

%----------------------------------------------------------------------------

%------------------------------------------------------------------------
% \subsection{Experiments on VIPeR}
%------------------------------------------------------------------------
\subsection{Experiments on MARS}
We also evaluate our method on video-based dataset. On this dataset, we employ two features as the baseline methods, LOMO and IDE. For each sequence, we first extract feature for each image, and use max pooling to combine all features into a fixed-length vector.
We set $k_1$ to 20, $k_2$ to 6, and $\lambda$ to 0.3 in this dataset. 
The performance of our method on different features and metrics are reported in Table~\ref{tabel:Mars-compare}.
As we can see, our re-ranking method consistently improves the rank-1 accuracy and mAP of the two different features. Results compared with average query expansion (AQE) \cite{chum2007total1} and contextual dissimilarity measure (CDM) \cite{jegou2007contextual} show our method outperforms them in both rank-1 accuracy and mAP.  
Moreover, our method can even improve the rank-1 accuracy and mAP in all cases while discriminative metrics are used. 
In particular, our method improves the rank-1 accuracy from 70.51$\%$ to 73.94$\%$ and the mAP from 55.12$\%$ to 68.45$\%$ for IDE (R) + XQDA. Experimental results demonstrate that our re-ranking method is also effective on video-based re-ID problem. We believe that results of this problem will be further improved by combining more sophisticated feature model with our method. 

%-----------------------------------------------------------------
\begin{table}
\footnotesize
\begin{center}
\caption{\label{tabel:Mars-compare} Comparison among various methods with our re-ranking approach on the  MARS dataset.  
%The proposed method consistently improves the performance of person re-identification, especially for mAP.
}\vspace{.03in}
\newcolumntype{C}{>{\centering\arraybackslash}X}%
\newcolumntype{R}{>{\raggedleft\arraybackslash}X}%
\begin{tabularx}{\linewidth}{ l|C|C }
\hline
Method & Rank 1 & mAP \\
\hline
\hline
% LOMO \cite{liao2015lomo} & 19.04 & 9.80  \\ 
% LOMO  + Re-ranking  & 17.22	 & 12.07   \\
LOMO + KISSME & 30.86	 & 15.36  \\
LOMO  + KISSME + Ours & 31.31  & 22.38  \\
LOMO + XQDA \cite{liao2015lomo} & 31.82	 & 17.00 \\
LOMO   + XQDA + Ours & 33.99  & 23.20  \\
\hline
IDE (C) \cite{prw}  & 61.72	 & 41.17  \\
IDE (C) + AQE \cite{chum2007total1}   & 61.83	 & 47.02 \\
IDE (C) + CDM \cite{jegou2007contextual}  & 62.05	 & 44.23 \\
IDE (C)  + Ours&  62.78 	 & 51.47   \\
IDE (C) +  KISSME & 65.25	 & 44.83  \\
IDE (C)   + KISSME + Ours & 66.87 	 & 56.18   \\
IDE (C) + XQDA & 65.05	 & 46.87  \\
IDE (C)   + XQDA + Ours & 67.78 	 & 57.98   \\
\hline
IDE (R) \cite{prw} & 62.73	 & 44.07 \\
IDE (R) + AQE \cite{chum2007total1} & 63.74	 & 49.14 \\
IDE (R) + CDM \cite{jegou2007contextual} & 64.11	 & 47.68 \\
IDE (R)  + Ours &  65.61 & 57.94  \\
IDE (R) + KISSME  & 70.35	 & 53.27  \\
IDE (R) + KISSME + Ours&  72.32 & 67.29 \\
IDE (R) + XQDA  & 70.51	 & 55.12  \\
\textbf{IDE (R) + XQDA + Ours}& \textbf{73.94}  &\textbf{68.45} \\
\hline
\end{tabularx}
\end{center}
\end{table}
%----------------------------------------------------------------------------

%-----------------------------------------------------------------
\begin{table}
\footnotesize
\begin{center}
\caption{\label{tabel:PRW-compare} Comparison among various methods with our re-ranking approach on the PRW dataset. 
%The proposed method consistently improves the performance of person re-identification, especially for mAP.
}\vspace{.03in}
\newcolumntype{C}{>{\centering\arraybackslash}X}%
\newcolumntype{R}{>{\raggedleft\arraybackslash}X}%
\begin{tabularx}{\linewidth}{ l|C|C }
\hline
Method & Rank 1 & mAP \\
\hline
\hline
LOMO + XQDA \cite{liao2015lomo} & 34.91 & 13.43  \\ 
LOMO  + XQDA + Ours  & 37.14 &19.22  \\
\hline
IDE (C) \cite{prw}  & 51.03	 & 25.09 \\
\textbf{IDE (C)  + Ours}&  \textbf{52.54}  & \textbf{31.51}   \\
\hline
\end{tabularx}
\end{center}
\end{table}
%----------------------------------------------------------------------------

%------------------------------------------------------------------------
\subsection{Experiments on PRW}
We also evaluate our method on the end-to-end re-ID dataset. This dataset is more challenging than image-based and video-based datasets, since it requires to detect person from a raw image and identify the correct person from the detected galleries. Following \cite{prw}, we first use DPM to detect candidate bounding boxes of persons on a large raw image, and then query on the detected bounding boxes. We use LOMO and IDE to extract features for each bounding box, and take these two methods as baselines. We set $k_1$ to 20, $k_2$ to 6, and $\lambda$ to 0.3. Experiment results are shown in Table~\ref{tabel:PRW-compare}. It can be seen that, our method consistently improves the rank-1 accuracy and mAP of both LOMO and IDE feature, demonstrating that our method is effective on end-to-end re-ID task. %Note that our method can be implemented to  ranking result of any feature or metric. 

\subsection{Parameters Analysis}
%-------------- 
The parameters of our method are analyzed in this subsection. The baseline methods are LOMO \cite{liao2015lomo} and IDE \cite{prw} trained on CaffeNet. We evaluate the influence of $k_1$, $k_2$, and $\lambda$ on rank-1 accuracy and mAP on the Market-1501 dataset. To conduct experimental analyses, we randomly split the original training set into training and validation sets, with 425 and 200 identities respectively.
%

\begin{figure}[!t]
\centering
   \subfigure{ \includegraphics[width=.24\linewidth]{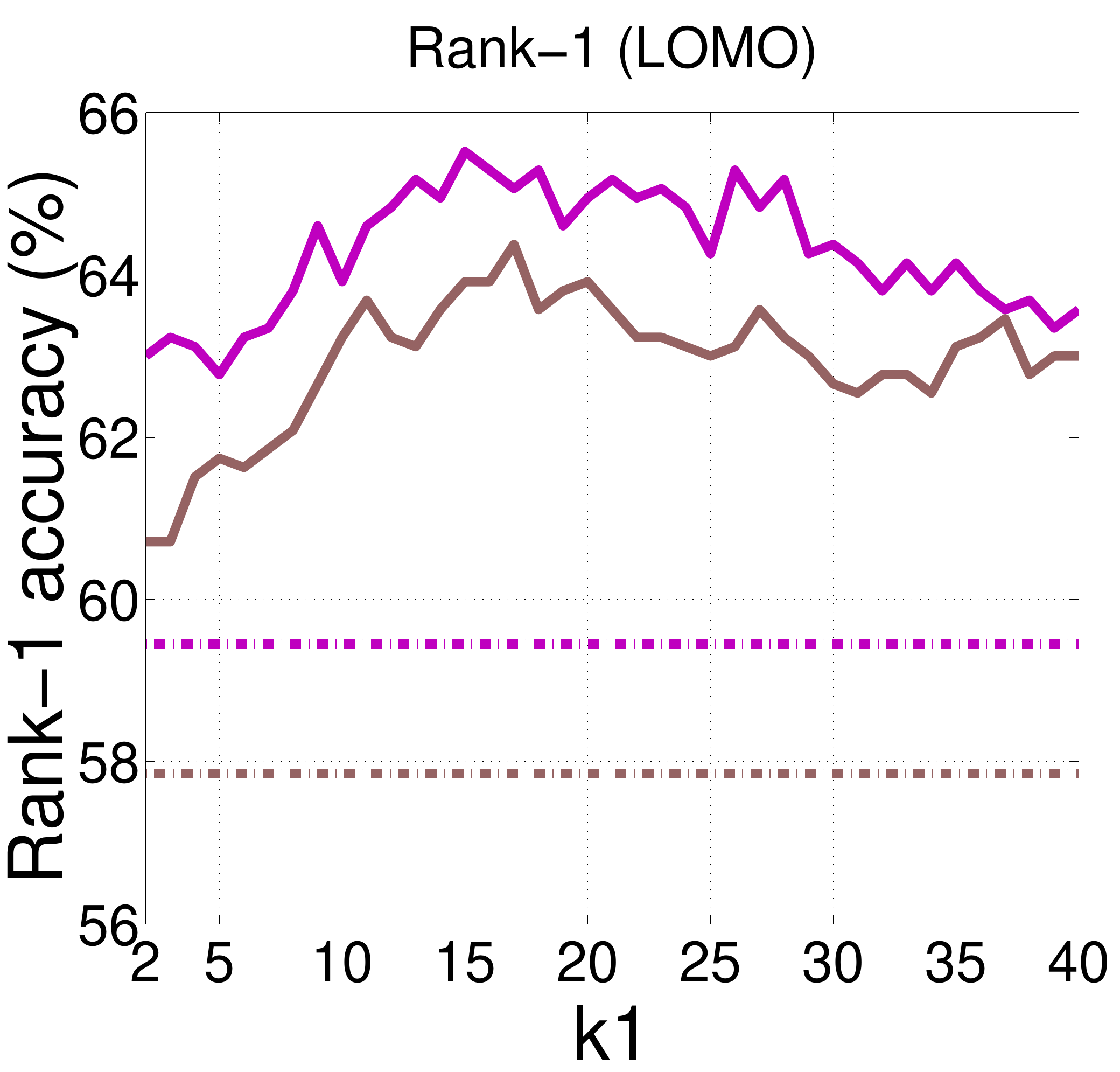}}\hspace{-.05in}
          \subfigure{ \includegraphics[width=.24\linewidth]{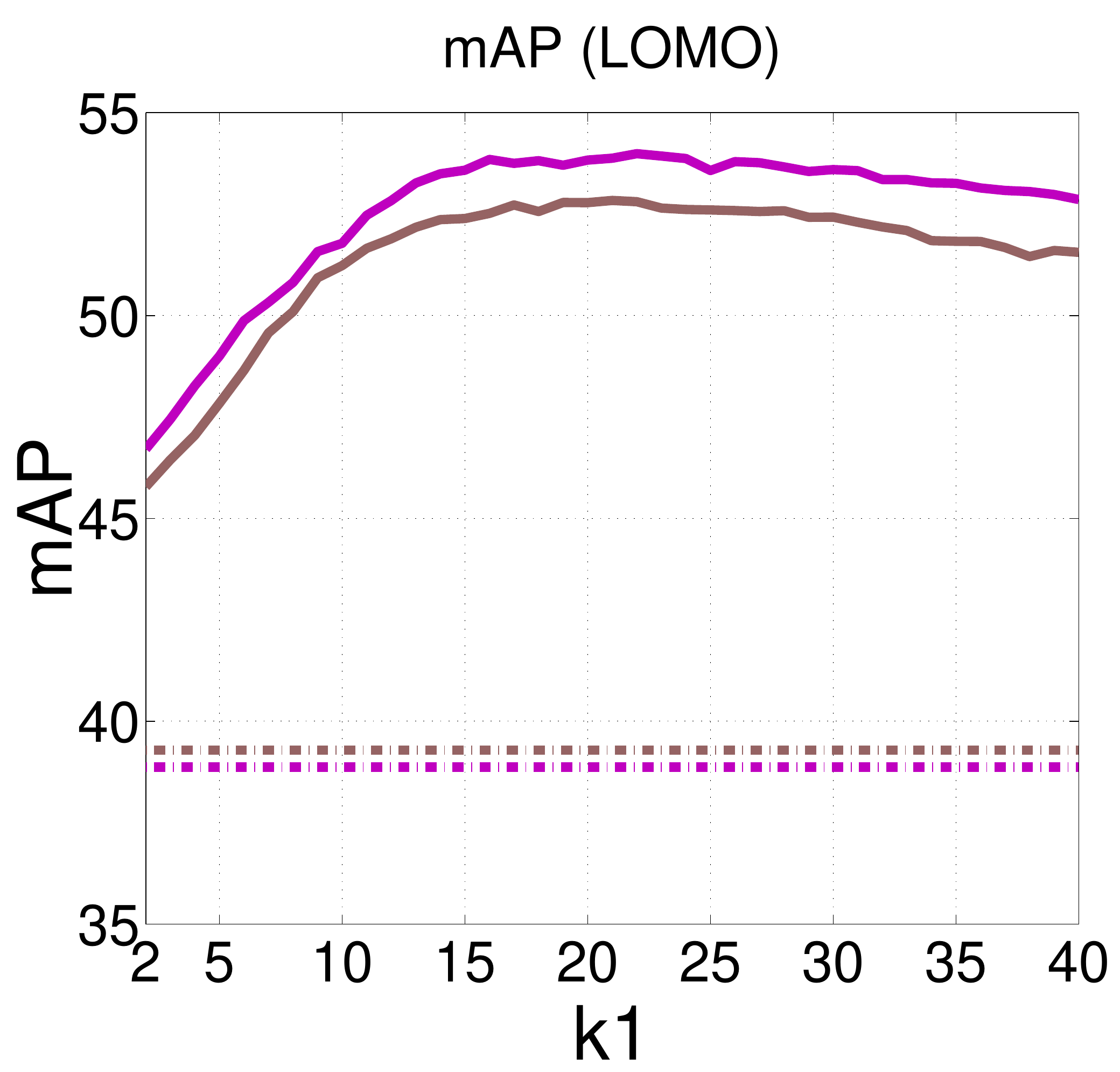}}\hspace{-.05in}
     \subfigure{ \includegraphics[width=.24\linewidth]{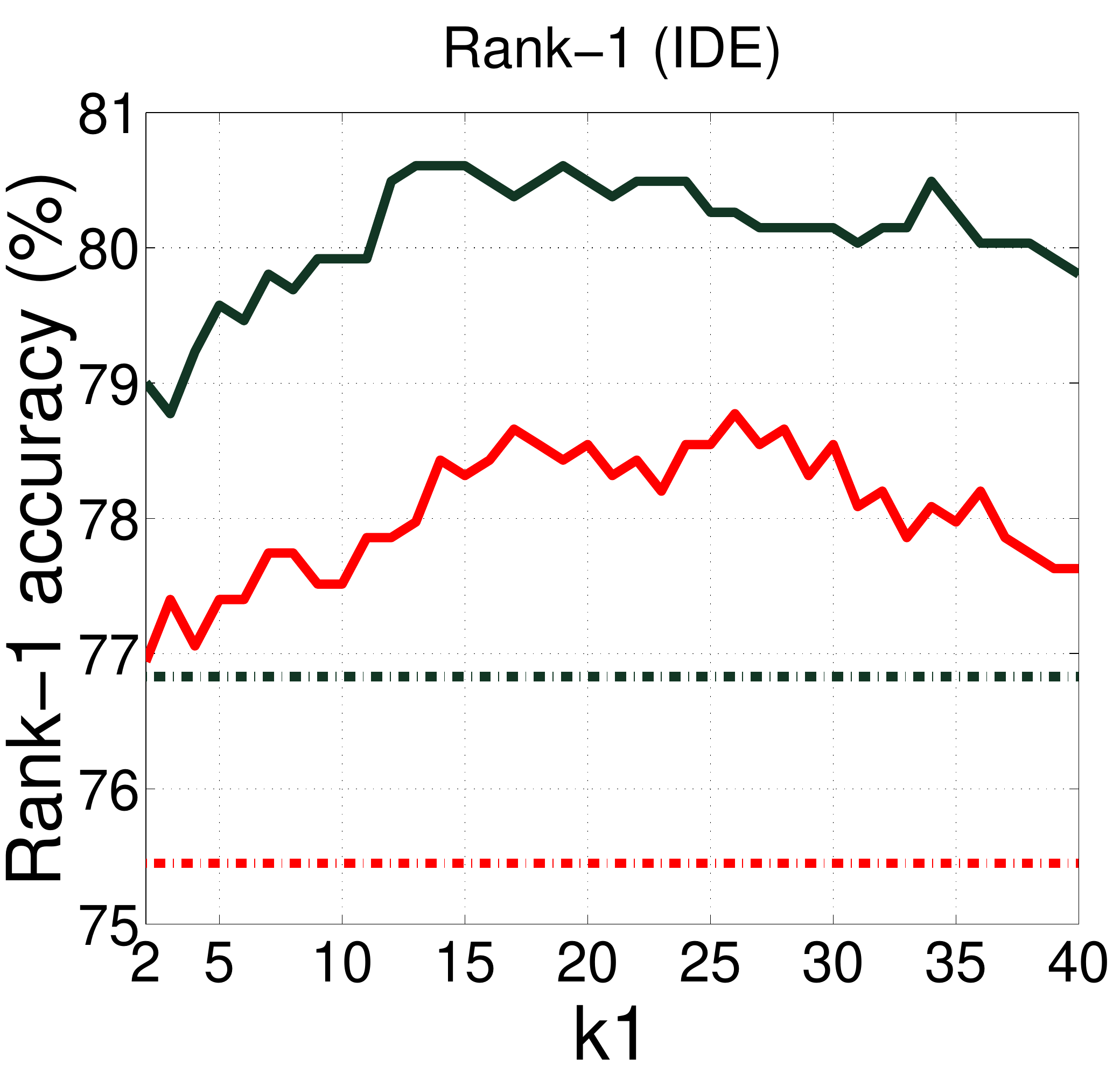}}\hspace{-.05in}
         \subfigure{ \includegraphics[width=.24\linewidth]{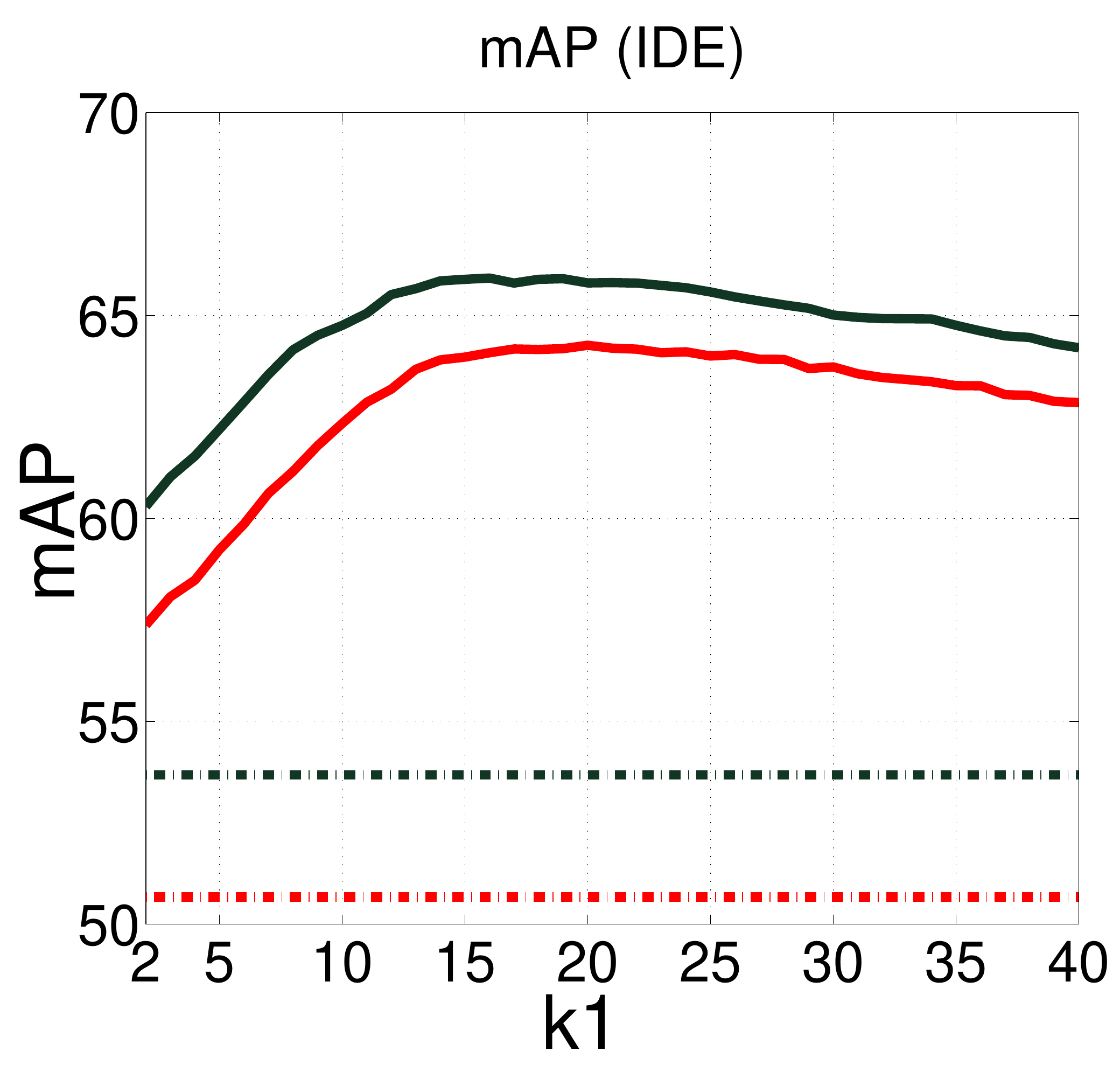}} \\\vspace{-.08in}
       \subfigure{
      \includegraphics[width=.98\linewidth]{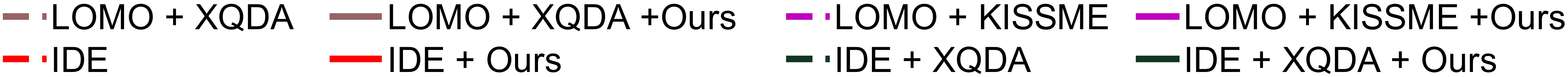}}%\vspace{-.05in}
\caption{The impact of the parameter $k_1$ on re-ID performance on the Market-1501 dataset. We fix the $k_2$ at 6 and $\lambda$ at 0.3.}
%\vspace{-.1in}
\label{fig:parameters_k1}
\vspace{-.1in}
\end{figure}

%---------------k1
Fig.~\ref{fig:parameters_k1} shows the impact of the size of k-reciprocal neighbors set on rank-1 accuracy and mAP. It can be seen that, our method consistently outperforms the baselines both on the rank-1 accuracy and mAP with various values of $k_1$. The mAP first increases with the growth of $k_1$, and then begins a slow decline after $k_1$ surpasses a threshold. Similarly, as $k_1$ grows, the rank-1 accuracy first rises with fluctuations; and after arriving at the optimal point around $k_1=20$, it starts to drop. With a too large value of $k_1$, there will be more false matches included in the k-reciprocal set, resulting in a decline in performance.

%---------------------------------------------------------------
%--------------------------------

%---------------------------------------------------------------
\begin{figure}[!t]
\centering
  \subfigure{ \includegraphics[width=.24\linewidth]{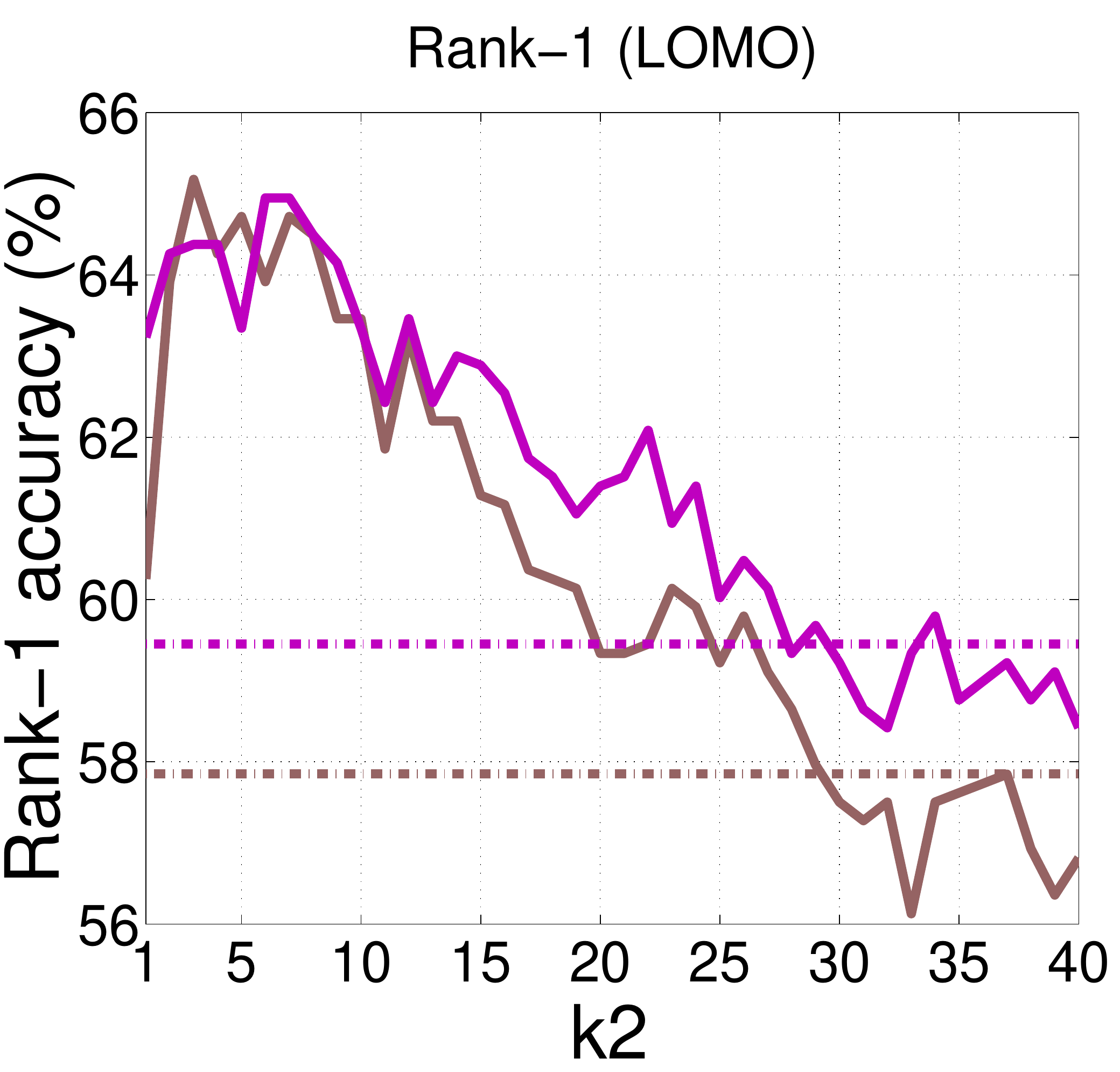}}\hspace{-.05in}
      \subfigure{ \includegraphics[width=.24\linewidth]{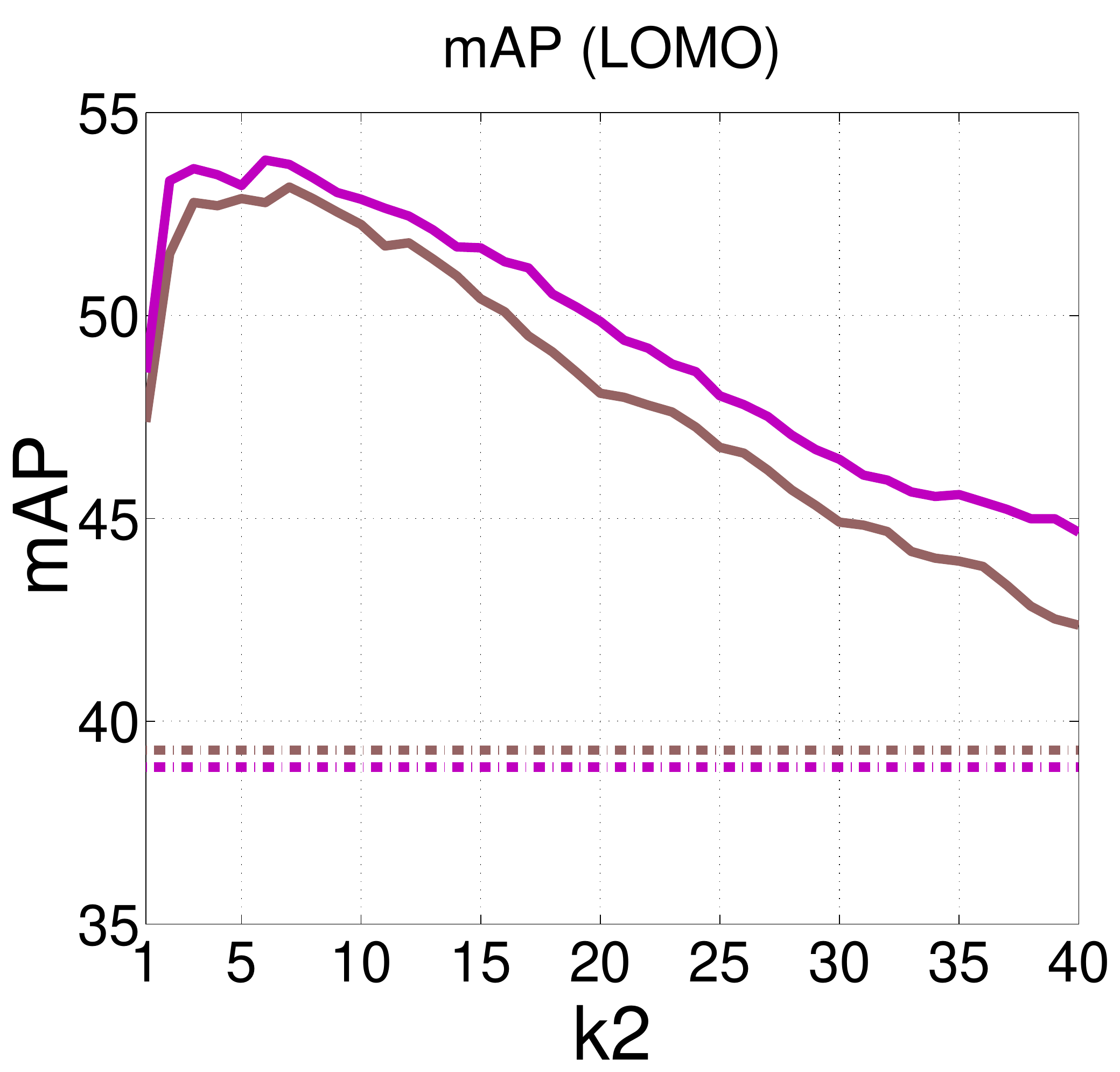}}\hspace{-.05in}
  \subfigure{ \includegraphics[width=.24\linewidth]{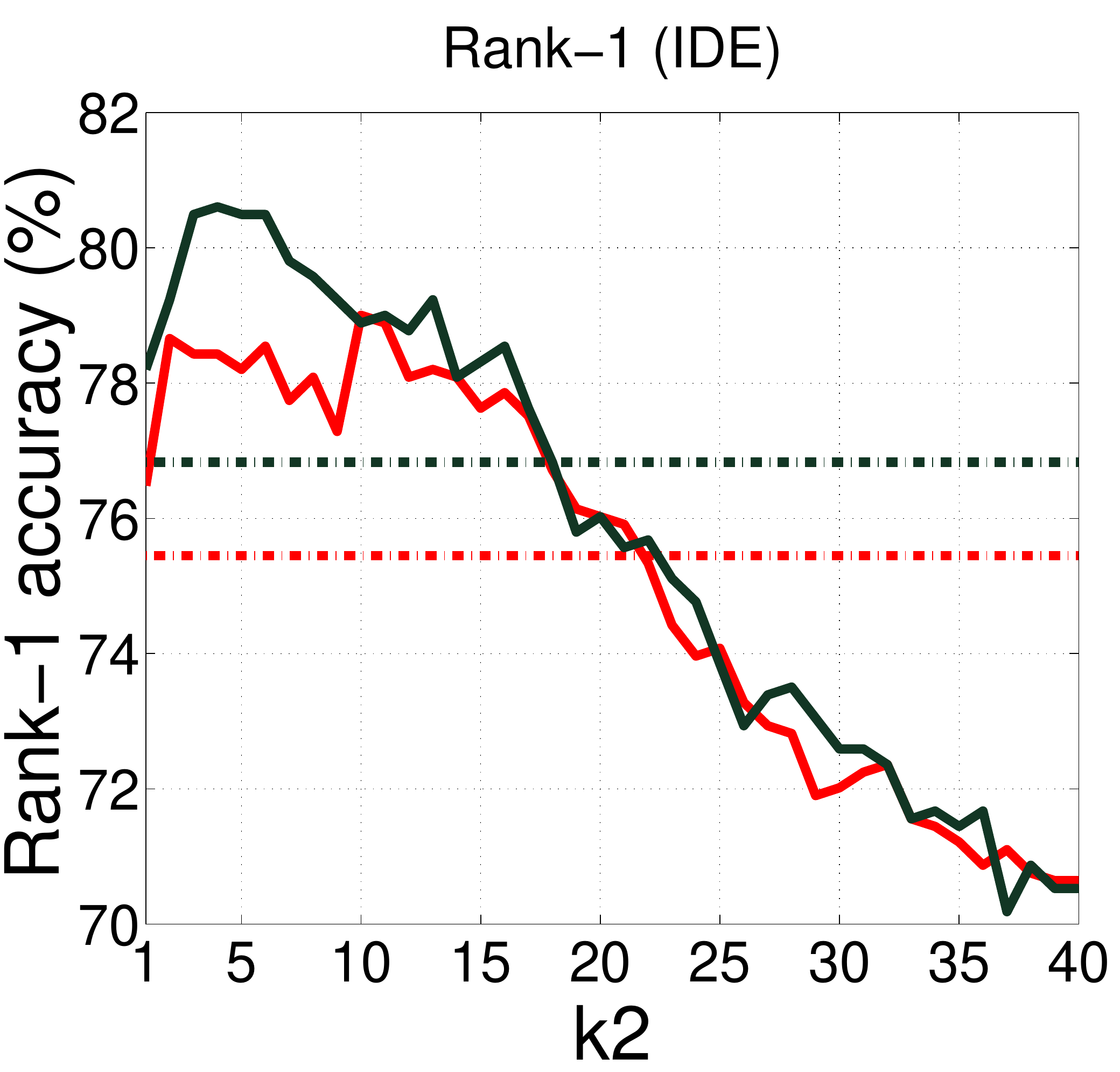}}\hspace{-.05in}
      \subfigure{ \includegraphics[width=.24\linewidth]{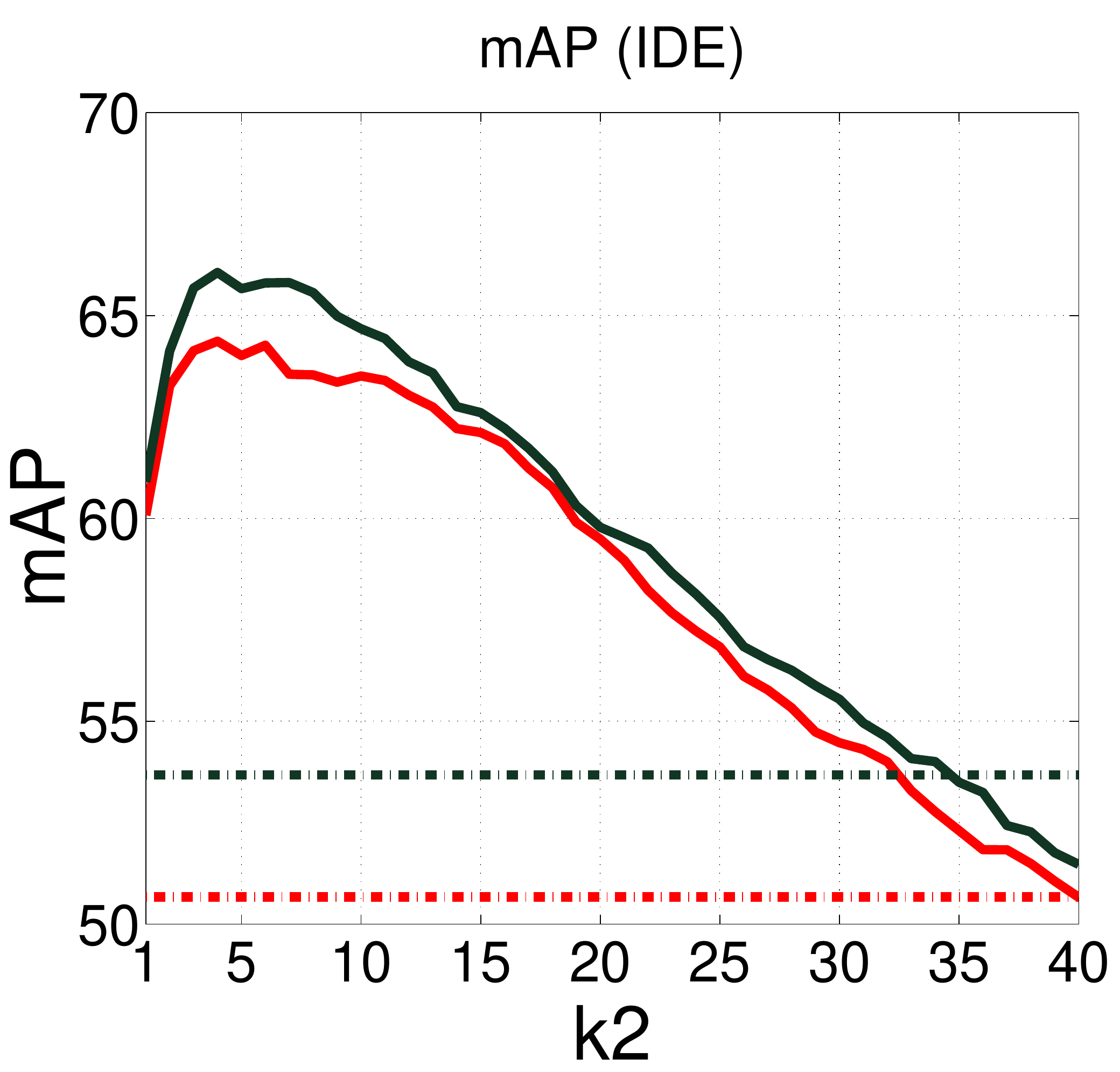}} \\\vspace{-.08in}
       \subfigure{
      \includegraphics[width=.98\linewidth]{image/legend_combine_trainval}}%\vspace{-.05in}
\caption{The impact of the parameter $k_2$ on re-ID performance on the Market-1501 dataset. We fix the $k_1$ at 20 and $\lambda$ at 0.3.}
%\vspace{-.1in}
\label{fig:parameters_k2}
\end{figure}
%--------------------------------

%---------------k2
The impact of $k_2$ are shown in Fig.~\ref{fig:parameters_k2}. When $k_2$ is equal to 1, the local query expansion is not considered.  Obviously, the performance grows as $k_2$ increases in a reasonable range. Notice that, assigning a much too large value to $k_2$ reduces the performance. Since it may lead to exponentially containing false matches in local query expansion, which undoubtedly harm the feature and thus the performance. As a matter of fact, the local query expansion is very beneficial for further enhancing the performance when setting an appropriate value to $k_2$.

\begin{figure}[!t]
\centering
\subfigure{ \includegraphics[width=.24\linewidth]{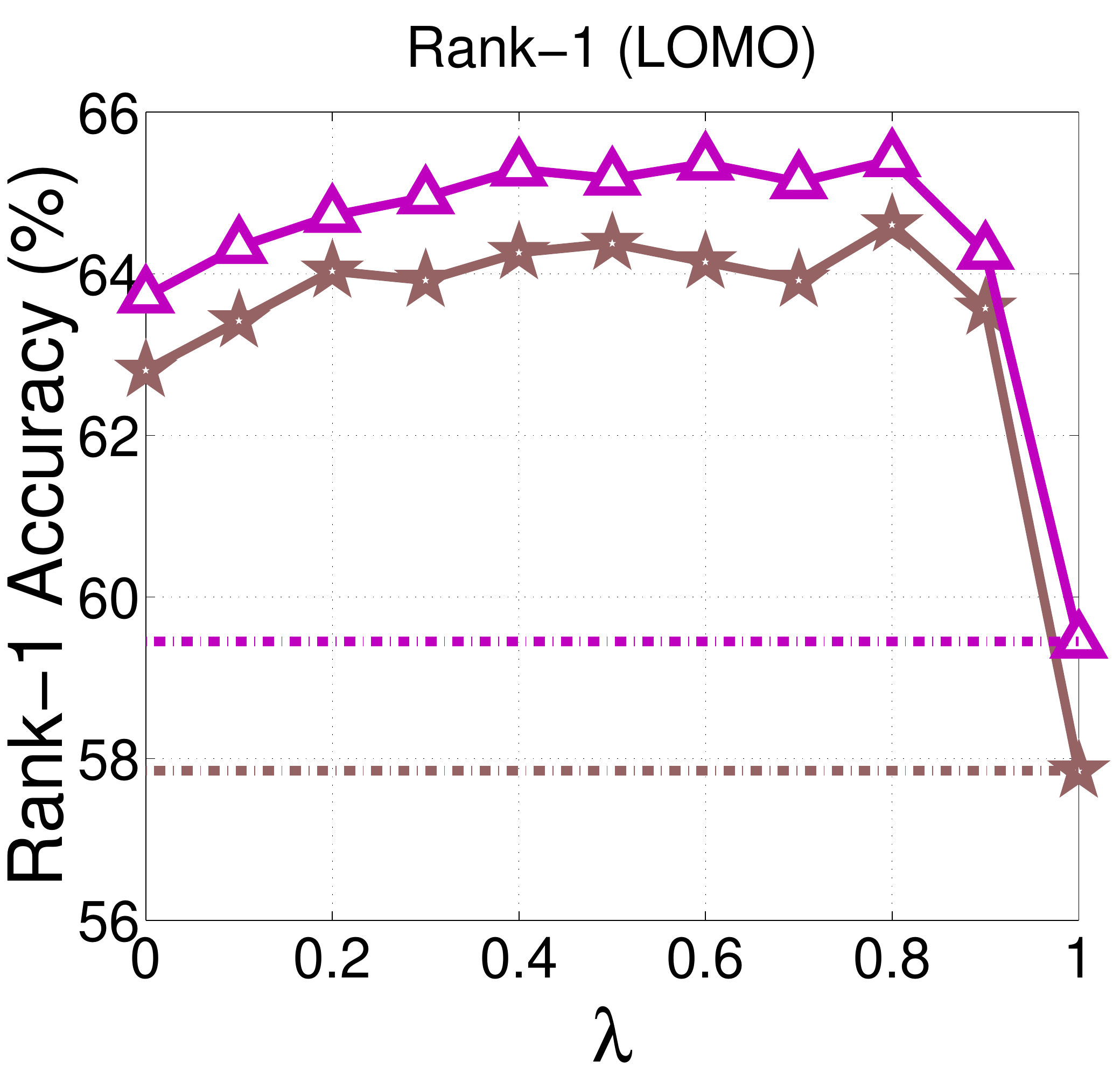}}\hspace{-.05in}
      \subfigure{ \includegraphics[width=.24\linewidth]{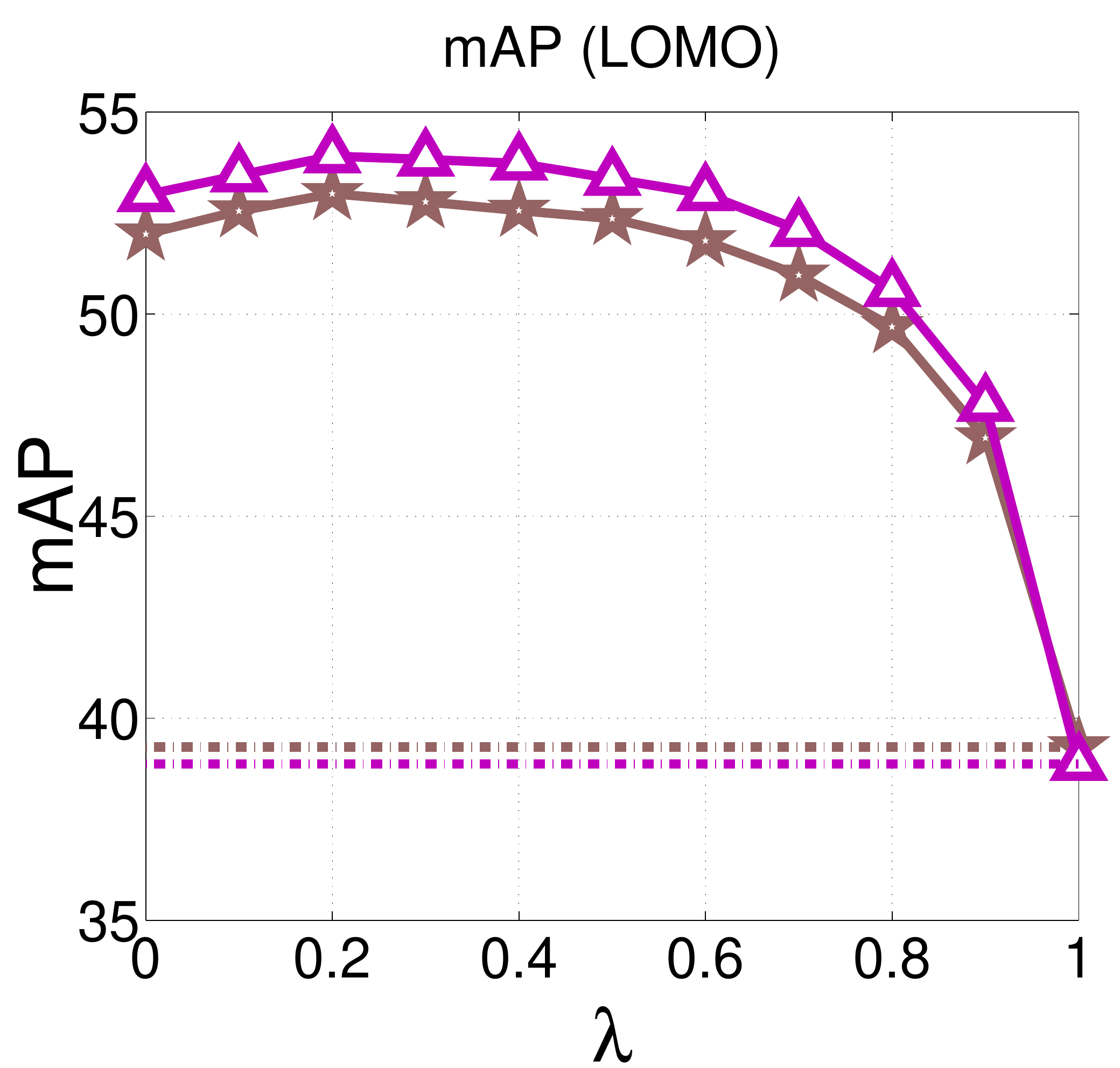}}\hspace{-.05in}
  \subfigure{ \includegraphics[width=.24\linewidth]{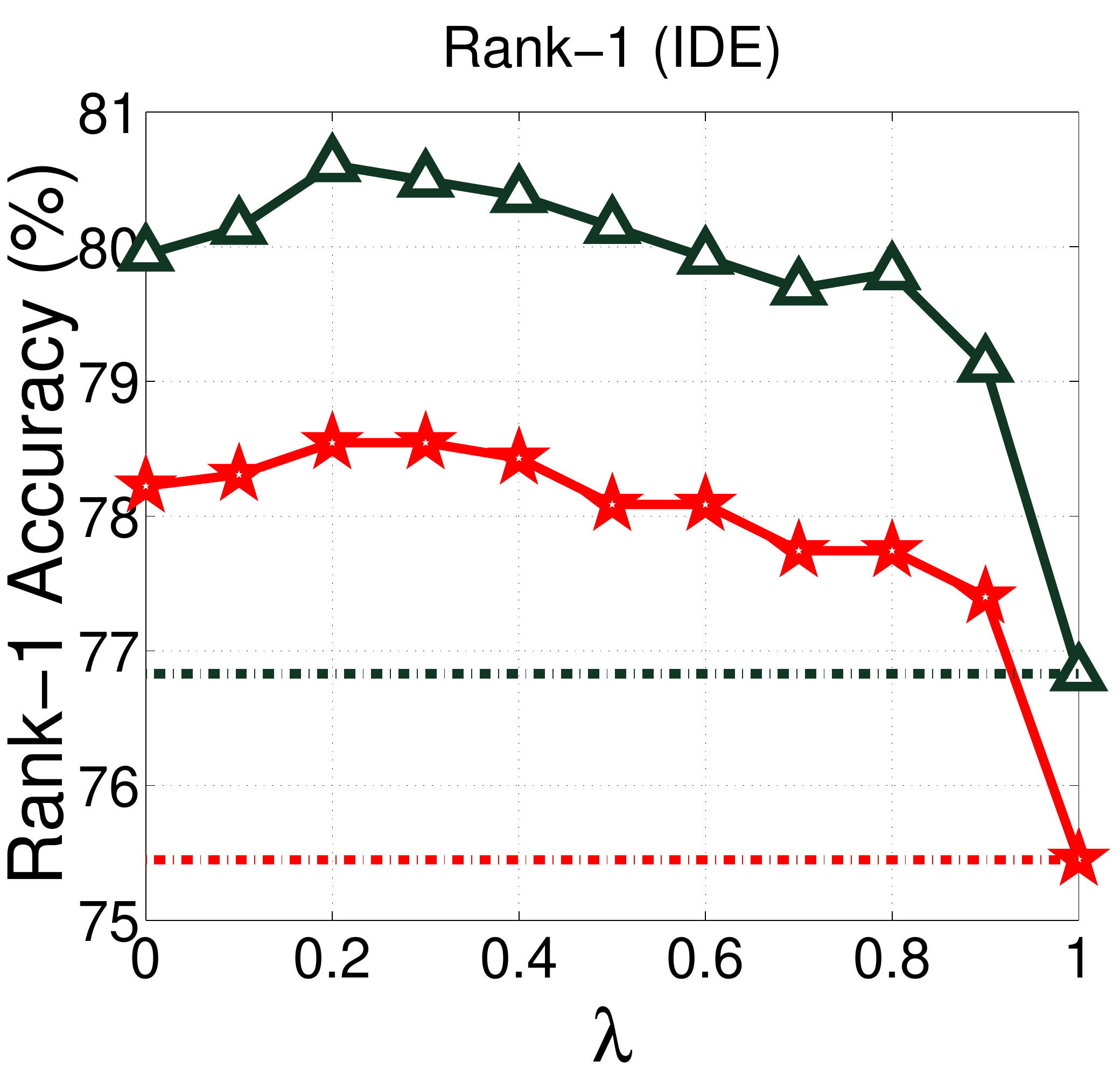}}\hspace{-.05in}
      \subfigure{ \includegraphics[width=.24\linewidth]{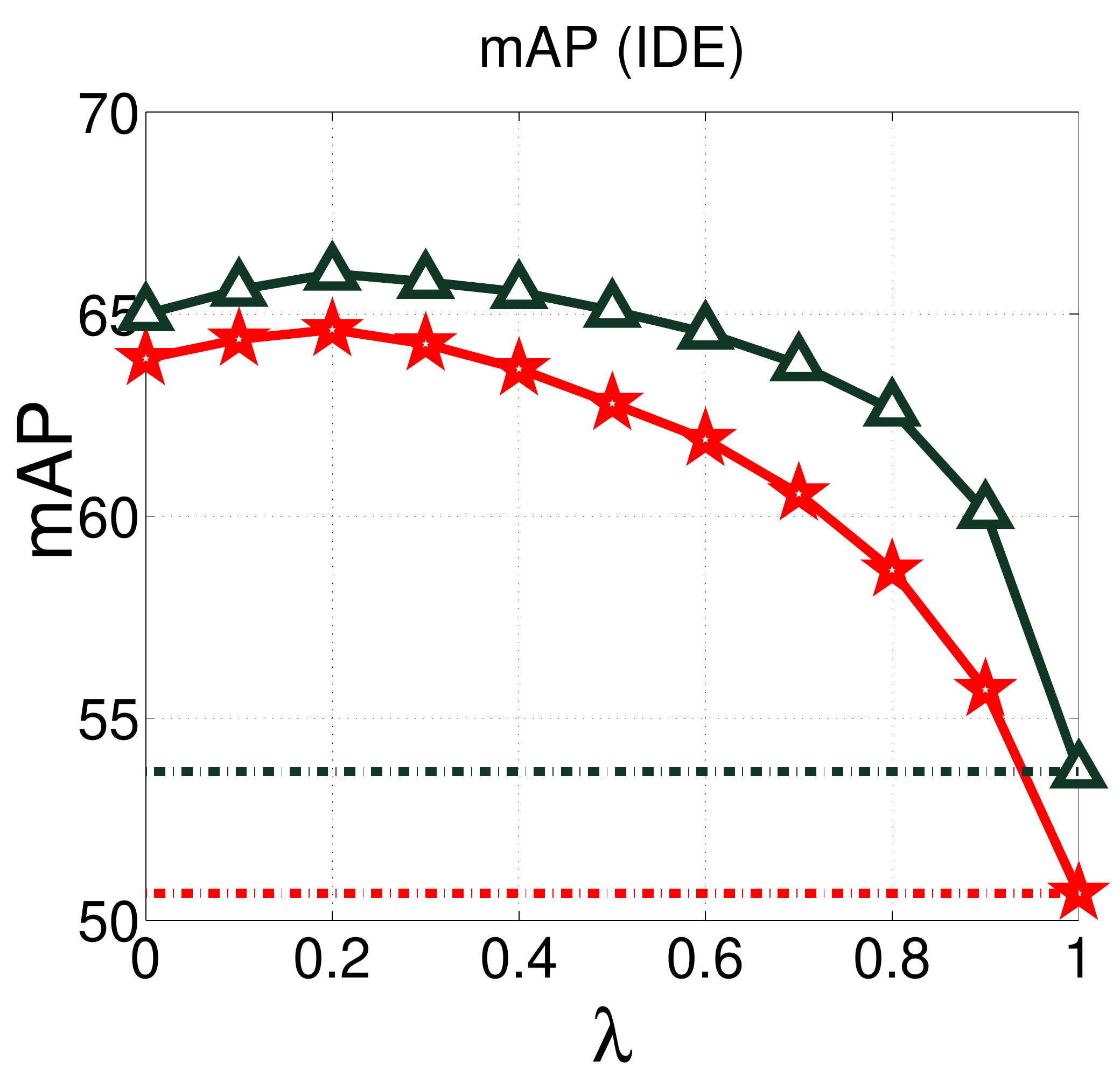}}\\\vspace{-.08in}
       \subfigure{
      \includegraphics[width=.98\linewidth]{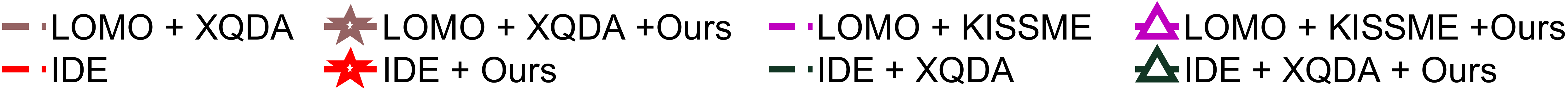}}%\vspace{-.05in}
\caption{The impact of the parameter $\lambda$ on re-ID performance on the Market-1501 dataset. We fix the $k_1$ at 20 and $k_2$ at 6.}
%\vspace{-.1in}
\label{fig:parameters_w}
\end{figure}

%---------------lambda
The impact of the parameter $\lambda$ is shown in Fig.~\ref{fig:parameters_w}.
Notice that, when $\lambda$ is set to 0, we only consider the Jaccard distance  as the final distance; in contrast, when  $\lambda$ equal to 1, the Jaccard distance is left out, and the result is exactly the baseline result obtained using pure original distance. It can be observed that when only Jaccard distance is considered, our method consistently outperforms the baseline. This demonstrates that the proposed Jaccard distance is effective for re-ranking. Moreover, when simultaneously considering both the original distance and the Jaccard distance, the performance obtains a further improvement when the value of $\lambda$ is around 0.3, demonstrating that the original distance is also important for re-ranking.

\begin{figure}[!t]
\centering
\includegraphics[width=\linewidth]{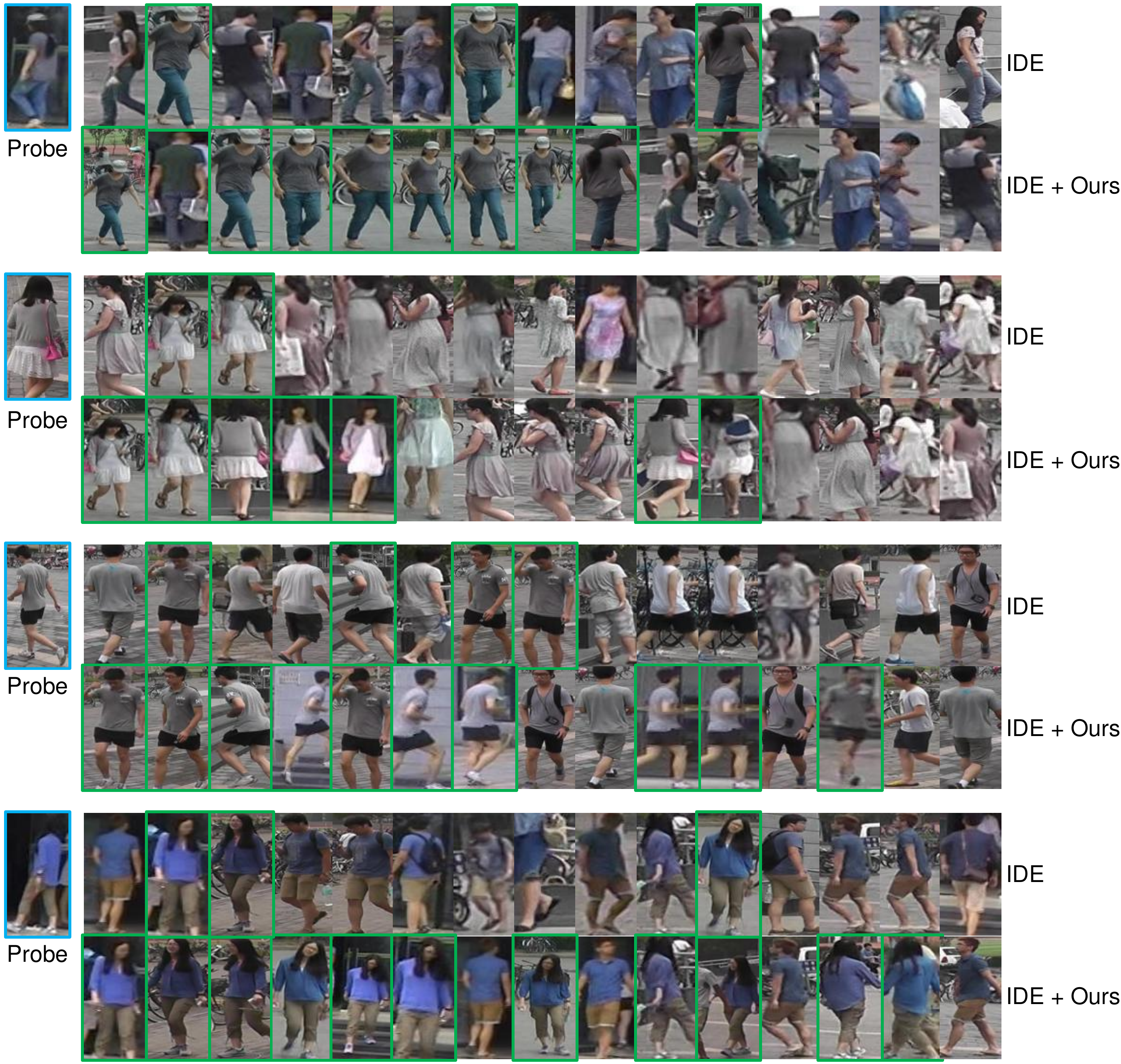}
\caption{Example results of four probes on the Market-1501 dataset. For each probe, the first row and the second correspond to the ranking results produced by IDE and IDE + Ours, respectively. Person surrounded by green box denotes the same person as the probe.}
\label{fig:sample_results}
\end{figure}

In Fig.~\ref{fig:sample_results}, four example results are shown. The proposed method, IDE + Ours, effectively ranks more true persons in the top of ranking list which are missed in the ranking list of IDE.

%------------------------------------------------------------------------
\section{Conclusion}
In this paper, we address the re-ranking problem in person re-identification (re-ID). We propose a $k$-reciprocal feature by encoding the k-reciprocal nearest neighbors into a single vector, thus the re-ranking process can be readily performed by vector comparison. To capture the similarity relationships from similar samples, the local expansion query is proposed to obtain a more robust $k$-reciprocal feature. The final distance based on the combination of the original distance and Jaccard distance produces effective improvement of the re-ID performance on several large-scale datasets. It is worth mentioning that our approach is fully automatic and unsupervised, and can be easily implemented to any ranking result.

%-----------\lambda----------------------------------------------------
%--------------------------------

%-------------------fig:sample result------------
%--------------------------------

%------------------------------------------------------------------------

 \section{Acknowledgements}
We thank Wenjing Li and Mingyi Lei for helpful discussions and encouragement. This work is supported by the Nature Science Foundation of China (No. 61572409, No.61402386 \& No. 61571188), Fujian Province 2011 Collaborative Innovation Center of TCM Health Management and Collaborative Innovation Center of Chinese Oolong Tea Industry—Collaborative Innovation Center (2011) of Fujian Province.

{\small
\bibliographystyle{ieee}
\bibliography{person}
}

\end{document}